\documentclass[journal]{IEEEtran}

\usepackage[usenames,dvipsnames,svgnames,table]{xcolor}

\usepackage{graphics} % for pdf, bitmapped graphics files
\usepackage{amsmath} % assumes amsmath package installed
\usepackage{amssymb}  % assumes amsmath package installed
\usepackage{pgfplots}

\definecolor{plotcolor1}{RGB}{249,10,10}
\definecolor{plotcolor2}{RGB}{47,154,255}
\definecolor{plotcolor3}{RGB}{32,187,32}
\definecolor{plotcolor4}{RGB}{238,136,16}
\definecolor{plotcolor5}{RGB}{10,10,249}
\pgfplotscreateplotcyclelist{theplotstyles}{%
solid,color=plotcolor1,thick\\%
solid,color=plotcolor2,thick\\%
solid,color=plotcolor3,thick\\%
solid,color=plotcolor4,thick\\%
solid,color=plotcolor5,thick\\%
}
\pgfplotscreateplotcyclelist{theplotstyleswithmarkers}{%
solid,color=plotcolor1,thick,mark=o,mark options={solid}\\%
solid,color=plotcolor2,thick,mark=o,mark options={solid}\\%
}

\usepackage{algcompatible}
\usepackage{algorithm, float}
\usepackage[noend]{algpseudocode}
\algnewcommand{\LineComment}[1]{\State \(\triangleright\) #1}
\makeatletter
\def\BState{\State\hskip-\ALG@thistlm}
\makeatother

\usepackage[caption=false,font=footnotesize]{subfig}
\usepackage{fixltx2e}
\usepackage{todonotes}
\usepackage{cite}
\usepackage{color}
\PassOptionsToPackage{dvipsnames}{xcolor}

\bstctlcite{IEEEexample:BSTcontrol}

\usepackage{booktabs}

\title{\LARGE \bf
A Random Finite Set Approach for Dynamic Occupancy Grid Maps with Real-Time Application }

\author{Dominik Nuss$^{1}$, Stephan Reuter$^{2}$, 
Markus Thom$^{2}$,  
Ting Yuan$^{1}$, \\
Gunther Krehl$^{1}$, Michael Maile$^{1}$,  Axel Gern$^{1}$, Klaus Dietmayer$^{2}$ % <-this % stops a space
\thanks{$^{1}$Mercedes-Benz Research \& Development North America, Inc., USA  
        {\tt\footnotesize dominik.s.nuss@daimler.com} or 
        {\tt\footnotesize firstname.lastname@daimler.com}, respectively}%
\thanks{$^{2}$Institute of Measurement, Control, and Microtechnology, Ulm University, Germany
        {\tt\footnotesize firstname.lastname@uni-ulm.de}}%
\thanks{S. Reuter is supported by the German Research Foundation (DFG) within the Transregional Collaborative Research Center SFB/TRR 62 "Companion-Technology for Cognitive Technical Systems".}%
}

\begin{document}

\maketitle
\thispagestyle{empty}
\pagestyle{empty}

\begin{abstract}
Grid mapping is a well established approach for environment perception in robotic and automotive applications. 
%It is particularly suited for data fusion of range finding sensors like laser, radar or stereo camera data, since their measurements can directly be associated to grid cells. 
Early work suggests estimating the occupancy state of each grid cell in a robot's environment using a Bayesian filter to recursively combine new measurements with the current posterior state estimate of each grid cell. 
This filter is often referred to as binary Bayes filter (BBF). 
A basic assumption of classical occupancy grid maps is a stationary environment. 
Recent publications describe bottom-up approaches using particles to represent the dynamic state of a grid cell and outline prediction-update recursions in a heuristic manner.   
This paper defines the state of multiple grid cells as a random finite set, which allows to model the environment as a stochastic, dynamic system with multiple obstacles, observed by a stochastic measurement system. 
It motivates an original filter called the probability hypothesis density / multi-instance Bernoulli (PHD/MIB) filter in a top-down manner.  
The paper presents a real-time application serving as a fusion layer for laser and radar sensor data and describes in detail a highly efficient parallel particle filter implementation. 
A quantitative evaluation shows that parameters of the stochastic process model affect the filter results as theoretically expected and that appropriate process and observation models provide consistent state estimation results.
\end{abstract}

%%%%%%%%%%%%%%%%%%%%%%%%%%%%%%%%%%%%%%%%%%%%%%%%%%%%%%%%%%%%%%%%%%%%%%%%%%%%%%%%

% indices and similar
\newcommand{\transp}{^\text{\rm \textup{T}}}
\newcommand{\est}[1]{\hat{#1}}
\newcommand{\kpred}{{+}}
\newcommand{\kp}{{k+1}}
\newcommand{\ind}{{(i)}}

% vectors
\newcommand{\x}{x}
\newcommand{\xl}{\mathbf{x}}
\newcommand{\xk}{\x_k}
\newcommand{\xp}{\x_{k+1}}

\newcommand{\z}{z}
\newcommand{\zk}{\z_k}
\newcommand{\zp}{\z_{k+1}}
\newcommand{\zallk}{\z_{1:k}}
\newcommand{\zallp}{\z_{1:k+1}}

% functions
\newcommand{\trans}{f_{+}}
\newcommand{\likek}{g_k}
\newcommand{\likep}{g_{k+1}}
\newcommand{\postk}{p_k}
\newcommand{\postp}{p_{k+1}}
\newcommand{\predk}{p_{\kpred}}
\newcommand{\GaussDist}[3]{\mathcal{N}\left( #1; #2, #3\right)}
\newcommand{\birth}{b}
\newcommand{\bk}{b_k}
\newcommand{\bp}{b_{k+1}}
\newcommand{\bpred}{b_{\kpred}}

\newcommand{\postok}{p_{o,k}}
\newcommand{\postop}{p_{o,k+1}}
\newcommand{\predok}{p_{o,\kpred}}

\newcommand{\proposp}{q_\kp}
\newcommand{\birthp}{b_\kp}

% spaces
\newcommand{\stateSpace}{\mathbb{X}}
\newcommand{\measSpace}{\mathbb{Z}}
\newcommand{\labelSpace}{\mathbb{L}}

% RFS BERNOULLI STATE
\newcommand{\sX}{\text{X}}
\newcommand{\sXl}{\protect\ensuremath\mathbf{X}}
\newcommand{\sZ}{\text{Z}}
\newcommand{\spi}{\pi}
\newcommand{\spil}{\boldsymbol{\spi}}

\newcommand{\ci}{{(c)}}
\newcommand{\Si}{{(S)}}

\newcommand{\spipredb}{\spi^\ci_{\text{b},\kpred}}
\newcommand{\spipredp}{\spi^\ci_{\text{p},\kpred}}
\newcommand{\spipred}{\spi^\ci_{\kpred}}
\newcommand{\spipredraw}{\spi_{\kpred}}

\newcommand{\spip}{\spi^\ci_{k+1}}
\newcommand{\spipraw}{\spi_{k+1}}

\newcommand{\sXp}{\sX_{k+1}}
\newcommand{\sXk}{\sX_{k}}
\newcommand{\sZp}{\text{Z}_{k+1}}
\newcommand{\spik}{\spi^\ci_{k}}

\newcommand{\spikraw}{\spi_{k}}

% GRID CELL BERNOULLI
% persistent object existence probability r
\newcommand{\experskp}{r^\ci_{\text{p},\kp}}
\newcommand{\experskpn}{\overline{r}^\ci_{\text{p},\kp}}
\newcommand{\experspred}{r^\ci_{\text{p},\kpred}}
\newcommand{\experspredn}{\overline{r}^\ci_{\text{p},\kpred}}
\newcommand{\expersk}{r^\ci_{\text{p},k}}
\newcommand{\experskn}{\overline{r}^\ci_{\text{p},k}}
% new-born object existence probability r
\newcommand{\exbornkp}{r^\ci_{\text{b},\kp}}
\newcommand{\exbornkpn}{\overline{r}^\ci_{\text{b},\kp}}
\newcommand{\exbornpred}{r^\ci_{\text{b},\kpred}}
\newcommand{\exbornpredn}{\overline{r}^\ci_{\text{b},\kpred}}
\newcommand{\exbornk}{r^\ci_{\text{b},k}}
\newcommand{\exbornkn}{\overline{r}^\ci_{\text{b},k}}

% combined object existence probability r
\newcommand{\exkp}{r^\ci_{\kp}}
\newcommand{\exkpn}{\overline{r}^\ci_{\kp}}
\newcommand{\expred}{r^\ci_{\kpred}}
\newcommand{\expredn}{\overline{r}^\ci_{\kpred}}
\newcommand{\exk}{r^\ci_{k}}
\newcommand{\exkn}{\overline{r}^\ci_{k}}

% cell observation
\newcommand{\cellTP}{p^\ci_{\text{TP},\kp}}
\newcommand{\cellTPn}{\overline{p}^\ci_{\text{TP},\kp}}
\newcommand{\cellFP}{p^\ci_{\text{FP},\kp}}
\newcommand{\cellFPn}{\overline{p}^\ci_{\text{FP},\kp}}
\newcommand{\zpcell}{\zp^\ci}
\newcommand{\pA}{p^\ci_{\text{A},\kp}}
\newcommand{\pAn}{\overline{p}^\ci_{\text{A},\kp}}

\newcommand{\likecell}{g^\ci_{\kp}(\zp|\xp)}
\newcommand{\likecellpartstar}{g^\ci_{\kp}(\zp|\partstarpredcell)}
\newcommand{\likecellpart}{g^\ci_{\kp}(\zp|\partperspredcell)}

\newcommand{\likecellA}{g^\ci_{\text{A},\kp}(\zp|\xp)}
\newcommand{\likecellApartstar}{g^\ci_{\text{A},\kp}(\zp|\partstarpredcell)}

\newcommand{\likecellAs}{g^\ci_{\text{A},\kp}}

%multi-object likelihood
\newcommand{\likemulti}{\gamma^\ci_\kp}
\newcommand{\likemultiraw}{\gamma_\kp}

% predicted persistent distribution
\newcommand{\pdfperspred}{p^\ci_{\text{p},\kpred}}

% PHD
\newcommand{\D}{D}
\newcommand{\Dk}{\D_k}
\newcommand{\Dpred}{\D_{\kpred}}
\newcommand{\Dp}{\D_{\kp}}

\newcommand{\Dpredb}{{^\text{b}\!\D}_{\kpred}}
\newcommand{\Dpb}{{^\text{b}\!\D}_{\kp}}

\newcommand{\Dpredp}{\D_{\text{p},\kpred}}
\newcommand{\Dpp}{{^\text{p}\!\D}_{\kp}}

%TODO: only use the following for persistent and predicted PHDs
\newcommand{\perD}{{^\text{p}\!\D}}
\newcommand{\bD}{{^\text{b}\!\D}}

\newcommand{\pDk}{{^\text{p}\!\D_k}}
\newcommand{\bDk}{{^\text{b}\!\D_k}}

\newcommand{\pDp}{{^\text{p}\!\D}_{\kp}}
\newcommand{\bDp}{{^\text{b}\!\D}_{\kp}}

\newcommand{\pDpred}{{^\text{p}\!\D}_{\kpred}}
\newcommand{\bDpred}{{^\text{b}\!\D}_{\kpred}}

%particles
\newcommand{\xpers}{{^\text{p} \hspace{-0.21mm}  \x}}
\newcommand{\wpers}{{^\text{p}\! w}}

\newcommand{\xborn}{{^\text{b} \hspace{-0.21mm}  \x}}
\newcommand{\wborn}{{^\text{b}\! w}}

\newcommand{\numParticles}{\nu}

\newcommand{\partk}{\x_k^\ind}
\newcommand{\weightk}{w_k^\ind}

\newcommand{\partkcell}{\x_{k}^{(i,c)}}
\newcommand{\weightkcell}{w_{k}^{(i,c)}}

\newcommand{\partperspred}{\x_{\text{p},\kpred}^{\ind}}
\newcommand{\weightperspred}{w_{\text{p},\kpred}^{\ind}}

\newcommand{\partstarpredcell}{\x_{*,\kpred}^{(i,c)}}

\newcommand{\partperspredcell}{\x_{\text{p},\kpred}^{(i,c)}}
\newcommand{\weightperspredcell}{w_{\text{p},\kpred}^{(i,c)}}

\newcommand{\partbornpredcell}{\x_{\text{b},\kpred}^{(i,c)}}
\newcommand{\weightbornpredcell}{w_{\text{b},\kpred}^{(i,c)}}

\newcommand{\partperspcell}{\x_{\text{p},\kp}^{(i,c)}}
\newcommand{\weightperspcell}{w_{\text{p},\kp}^{(i,c)}}

\newcommand{\partbornpcell}{\x_{\text{b},\kp}^{(i,c)}}
\newcommand{\weightbornpcell}{w_{\text{b},\kp}^{(i,c)}}

\newcommand{\numParticlesKCell}{\nu_{k}^\ci}
\newcommand{\numParticlesPCell}{\nu_{\kp}^\ci}

\newcommand{\numParticlesPredCell}{\nu_{\text{p},\kpred}^\ci}
\newcommand{\numParticlesBornCell}{\nu_{\text{b},\kpred}^\ci}

% TODO: correct this typo:
\newcommand{\numParticlesPredCellp}{\nu_{\text{p},\kp}^\ci}
\newcommand{\numParticlesPersCellp}{\nu_{\text{p},\kp}^\ci}
\newcommand{\numParticlesBornCellp}{\nu_{\text{b},\kp}^\ci}

\newcommand{\weightunnormcell}{\widetilde{w}_{\kp}^{(i,c)}}
\newcommand{\weightunnormcellstar}{\widetilde{w}_{*,\kp}^{(i,c)}}

\newcommand{\weightpredcell}{w_{\kpred}^{(i,c)}}
\newcommand{\weightpredcellstar}{w_{*,\kpred}^{(i,c)}}

\newcommand{\weightunnormperscell}{\widetilde{w}_{\text{p},\kp}^{(i,c)}}
\newcommand{\weightunnormborncell}{\widetilde{w}_{\text{b},\kp}^{(i,c)}}

\newcommand{\partpcell}{\x_{\kp}^{(i,c)}}
\newcommand{\weightpcell}{w_{\kp}^{(i,c)}}
\newcommand{\weightpcellstar}{w_{*,\kp}^{(i,c)}}

\newcommand{\normempty}{\mu^\ci_{\{\emptyset\}}}
\newcommand{\normz}{\mu^\ci_{\{\z\}}}

\newcommand{\numParticlesNewTotal}{\nu_\text{b}}

%Dempster-Shafer Particle sets

\newcommand{\partasspcell}{\x_{\text{A},\kp}^{(i,c)}}
\newcommand{\weightasspcell}{w_{\text{A},\kp}^{(i,c)}}
\newcommand{\numParticlesAssCellp}{\nu_{\text{A},\kp}^\ci}

\newcommand{\partassnpcell}{\x_{\overline{\text{A}},\kp}^{(i,c)}}
\newcommand{\weightassnpcell}{w_{\overline{\text{A}},\kp}^{(i,c)}}
\newcommand{\numParticlesAssNCellp}{\nu_{\overline{\text{A}},\kp}^\ci}

\newcommand{\numParticlesNewCell}{\nu_{\text{b},\kp}^\ci}

%particle numbers
\newcommand{\nuk}{{\nu_k}}
\newcommand{\nup}{{\nu_\kp}}
\newcommand{\nupredb}{{^\text{b}\! \nu_{\kpred}}}
\newcommand{\nupb}{{^\text{b}\! \nu_{\kp}}}
\newcommand{\nupredp}{{^\text{p}\! \nu_{\kpred}}}
\newcommand{\nupp}{{^\text{p}\! \nu_{\kp}}}

%Expected object birth numbers
\newcommand{\Nkb}{{^\text{b}\! \est{N}_k}}
\newcommand{\Npb}{{^\text{b}\! \est{N}_\kp}}
\newcommand{\Npp}{{^\text{p}\! \est{N}_\kp}}

%Expected Cardinality
\newcommand{\expcar}{\est{N}}
\newcommand{\Ncppred}{{^\text{p}\! \est{N}^c_{{k+1|k}}}}

% probabilites
\newcommand{\pD}{{p_\text{D}}}
\newcommand{\pS}{{p_\text{S}}}
\newcommand{\pB}{{p_\text{B}}}

% Grid Map

\newcommand{\pc}{{p_\text{cl}}}
\newcommand{\pb}{{p_\text{b}}}

\newcommand{\occEvent}{{O}}
\newcommand{\freeEvent}{{F}}

\newcommand{\occState}{{o}}
\newcommand{\noccState}{{\overline{o}}}
\newcommand{\occEventk}{\occEvent_k}
\newcommand{\occEventp}{\occEvent_{k+1}}
\newcommand{\occStatek}{\occState_k}
\newcommand{\occStatep}{\occState_{k+1}}
\newcommand{\occStatepred}{\occState_{\kpred}}
\newcommand{\noccStatek}{\overline{\occState}_k}
\newcommand{\noccStatep}{\overline{\occState}_{k+1}}

\newcommand{\freeEventk}{\freeEvent_k}
\newcommand{\freeEventp}{\freeEvent_{k+1}}

% Dempster Shafer

\newcommand{\bbazp}{m_{\zp}}

\newcommand{\bbakcell}{m_k^\ci}
\newcommand{\bbapredcell}{m_{\text{p},\kpred}^\ci}
\newcommand{\bbazpcell}{m_{\zp}^\ci}
\newcommand{\bbakpcell}{m_\kp^\ci}

\newcommand{\bbaperskpcell}{m_{\text{p},\kp}^\ci}
\newcommand{\bbabornkpcell}{m_{\text{b},\kp}^\ci}

\newcommand{\massperskpcell}{\varrho_{\text{p},\kp}^\ci}
\newcommand{\massbornkpcell}{\varrho_{\text{b},\kp}^\ci}

% Statistical Moments
\newcommand{\vxmean}{\overline{v}^\ci_\text{x}}
\newcommand{\vxmeanS}{\overline{v}^\Si_\text{x}}
\newcommand{\vxmeanSqared}{{\overline{v}^2}^\ci_\text{x}}
\newcommand{\vxmeanSSquared}{{\overline{v}^2}^\Si_\text{x}}
\newcommand{\vymean}{\overline{v}^\ci_\text{y}}

\newcommand{\partvx}{v_{\text{x},\text{p},\kp}^{(i,c)}}
\newcommand{\partvy}{v_{\text{y},\text{p},\kp}^{(i,c)}}

\newcommand{\varx}{{{\sigma^2}^\ci_{v_\text{x}}}}
\newcommand{\varxS}{{{\sigma^2}^\Si_{v_\text{x}}}}
\newcommand{\stddevS}{{{\sigma}^\Si_{v_\text{x}}}}

\newcommand{\covarxy}{\sigma^\ci_{v_\text{x} v_\text{y}}}

\newcommand{\occStateJoint}{\occState_k^{1:C}}

\newcommand{\dynState}{\mathbf{x}}
\newcommand{\dynStateCell}{\dynState_k^{c}}
\newcommand{\dynStateCellp}{\dynState_{k+1}^{c}}
\newcommand{\dynStateJoint}{\dynState_k^{1:C}}
\newcommand{\dynSpace}{\mathbb{R}^4}

\newcommand{\CombCellState}{\mathbf{X}_k^c}

\newcommand{\meas}{\mathbf{z}}
\newcommand{\measAll}{\meas_{1:k}}
\newcommand{\meask}{\meas_{k}}
\newcommand{\measkm}{\meas_{1:k-1}}

\newcommand{\bm}{\mathbf}

\newcommand{\cardlikeck}{g^{N_c}_k}
\newcommand{\cardlikecp}{g^{N_c}_{k+1}}

\newcommand{\cardpredc}{{N_c}_{k+1}}

\newcommand*{\QEDB}{\hfill\ensuremath{\square}}%
%%%%%%%%%%%%%%%

\section{Introduction}
The beginning of grid mapping approaches took place in the field of robotics \cite{Elfes1989, Thrun2005}.
%Mobile robots, equipped with range finding sensors like ultrasonic devices needed a free-space map of their surroundings. 
A classic grid map divides the environment into single grid cells and estimates the occupancy probability for each cell. 
Since several measurements occur over time, the grid map combines these measurements with a Bayesian filter. A commonly used filter for this application is the binary Bayes filter, which combines measurements to estimate the binary state of a grid cell: free or occupied \cite{Dietmayer2015}. A restrictive assumption of the common binary Bayes filter application is that the environment is stationary. Furthermore, a common assumption of grid maps is the independence of individual grid cells which facilitates a fast implementation at the cost of approximation errors.

Today, grid maps are used in many automated vehicles \cite{Nuss2013}, \cite{Urmson2008}, \cite{Kunz2015}. Due to their explicit free-space estimation and their ability to represent arbitrarily shaped objects, grid maps are an important tool for collision avoidance. Moreover, the spatial grid structure provides a convenient fusion layer for data from different range finding sensors
\cite{Laugier2011, Nuss2014a}. In vehicle environment perception, the assumption of a stationary environment is obviously not fulfilled due to moving road users like vehicles or pedestrians. 

Recently, several approaches have been presented to combine grid mapping and multi-object tracking. A well-known example is an approach called simultaneous localization, mapping and moving object tracking (SLAMMOT) \cite{Wang2007}, which retains a grid map and multiple object tracks at the same time and assigns object detections either to the grid map or to a tracked object. Other publications suggest associating grid cells directly to object tracks \cite{Bouzouraa2010} or detecting object movement in multiple time frames of grid maps using a post-processing step \cite{Vu2009}. Further approaches combine grid mapping and object tracking in a modular way \cite{Nuss2014}, \cite{Vatavu2015}.

However, some of these approaches imply complicated environment perception architectures and are therefore not an appropriate choice for many applications. In 2006, Cou{\'e} et al. proposed the Bayesian occupancy filter (BOF)\cite{Coue2006} which uses a four-dimensional grid to estimate a two-dimensional environment. Here, two grid dimensions represent the spatial position and two grid dimensions represent the two-dimensional velocity of the obstacles. Thus, the BOF estimates object movement and explicitly considers it in its process model. The BOF motivated many applications \cite{Laugier2011}, \cite{Gindele2009}, but a problem is the high computational load caused by the large number of grid cells necessary to represent the environment appropriately.

An important improvement by Danescu et al. \cite{Danescu2010, Danescu2011} suggested to represent the dynamic state of a grid cell with particles resulting in a significant reduction of the computational load. In subsequent publications, N\`{e}gre et al. \cite{Negre2014} and Tanzmeister et al. \cite{Tanzmeister2014} independently proposed to represent only the dynamic part of a grid map with particles instead of all occupied grid cells. In 2015, Nuss et al. suggested to use the dynamic grid map as a fusion layer for laser and radar measurements \cite{Nuss2015}, which would improve the overall performance of the dynamic grid map, especially the separation between moving and static obstacles.

In summary, previous work on dynamic grid maps based on particles shows promising results. 
Unfortunately, the proposed filters 
%lack a sound theoretical foundation and instead  
%They describe heuristically motivated prediction and update steps for particles, which are similar to classical Bayesian filters. 
%The mentioned publications 
lack a stochastically rigorous definition of a multi-object state estimation problem. 
As such, they describe evolutionary algorithms (survival of the fittest) rather than Bayesian filters.

\subsection{Contributions of this Paper}

This paper models the dynamic state estimation of grid cells as a random finite set (RFS) problem. Finite set statistics (FISST) \cite{Mahler2007} provide a mathematical framework for the state estimation of multiple dynamic objects in a Bayesian sense. Well-known techniques from the field of FISST like the probability hypothesis density filter (PHD) \cite{Mahler2003} and the Bernoulli filter (BF) \cite{Ristic2013c} are applied to estimate the dynamic state of grid cells. The resulting filter is called probability hypothesis density / multi-instance Bernoulli (PHD/MIB) filter. Modeling the estimation problem of a dynamic grid map in the random finite set domain yields substantial advantages. It gives every filter parameter a physical meaning and allows a generic and stochastically rigorous filter design for various estimation problems. 

The key contributions of this paper are:
\begin{enumerate}
\item The definition of the dynamic state estimation of grid cells as an RFS problem and the derivation of the probability hypothesis density / multi-instance Bernoulli (PHD/MIB) filter, which takes into account the special form of measurement grids as they are common for grid mapping approaches.
\item The realization of the PHD/MIB filter with particles and an approximation in the Dempster-Shafer domain.
\item A detailed pseudo code description of a massively parallel, real-time capable approximation of the PHD/MIB filter.
\item Results of experiments with real-world data and evaluation of estimation error and consistency of the approximated PHD/MIB filter.
\end{enumerate}

\subsection{Paper Structure}

The remainder of this paper is structured as follows. Section \ref{sec:dgm} gives an overview of published dynamic grid mapping approaches. Section \ref{sec:rfs} outlines mathematical basics of random finite set statistics. The PHD/MIB filter is derived in Sect. \ref{sec:phdmib}. A particle-based realization is presented in Sect. \ref{sec:implementation} and approximated in the Dempster-Shafer domain in Sect. \ref{sec:ds_appr}. Section \ref{sec:parallel_implementation} provides a detailed description of a highly-efficient parallel implementation, followed by the evaluation in Sect. \ref{sec:evaluation}. Section \ref{sec:conclusion} presents the conclusion.

\section{Dynamic Grid Mapping: An overview}
\label{sec:dgm}
This section provides an overview of current static and dynamic grid mapping approaches and discusses their advantages and drawbacks.

\subsection{Static Grid Mapping}

\begin{figure}[t]
  \centering
  \includegraphics[width=\columnwidth]{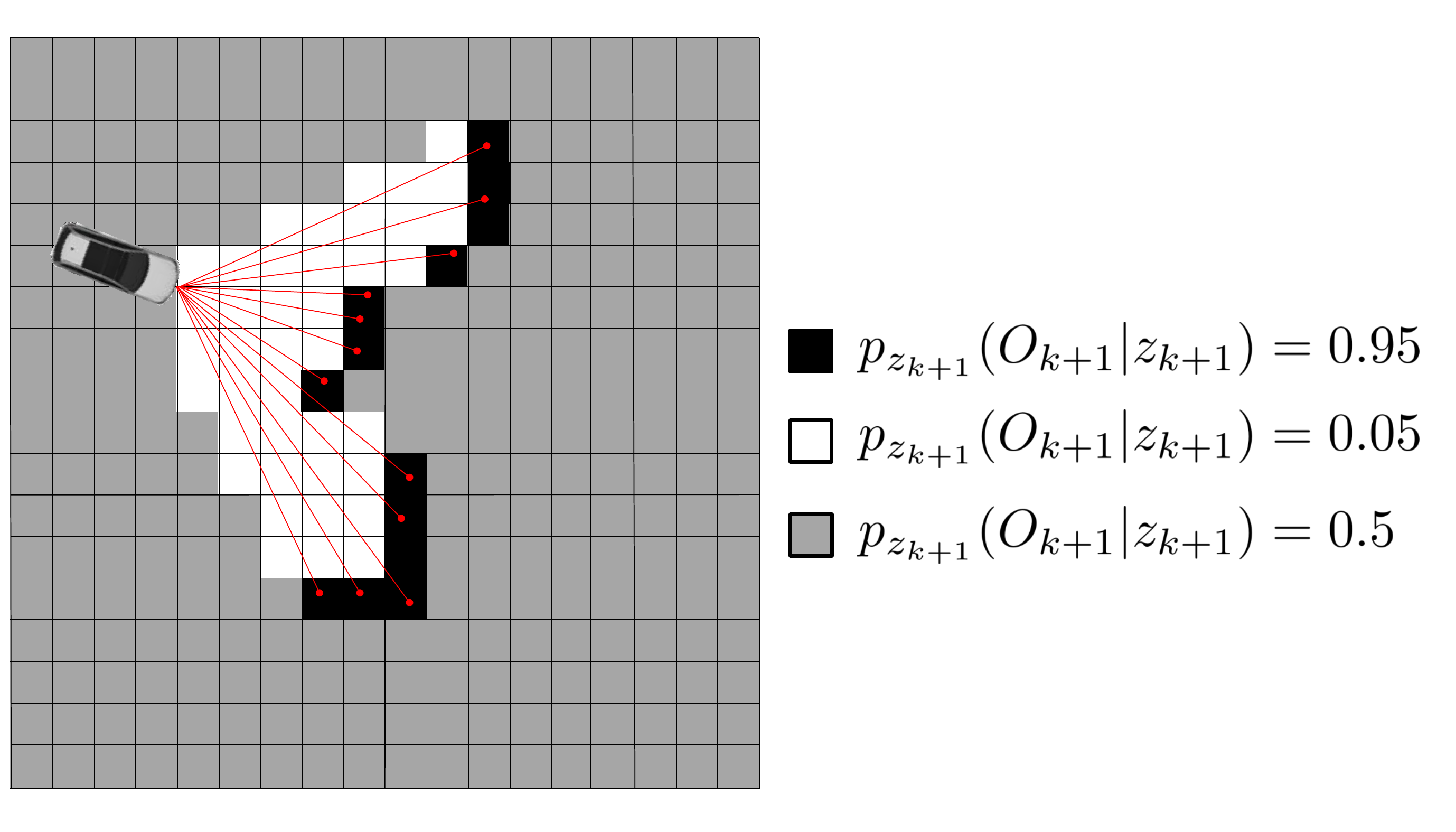}
	\caption{Occupancy probabilities of two-dimensional grid cells, reasoning on a multi-beam laser range measurement. Grid cells with a high probability of being occupied are colored black, free grid cells are marked with white color. Grid cells with an unknown state (same probability for both occupied and free) are displayed in gray color.}
	\label{fig:measGrid}
\end{figure}

Classic occupancy grid maps divide the space into single grid cells and estimate the occupancy probability of each grid cell \cite{Elfes1989,Thrun2005, Dietmayer2015}. 
A binary grid cell state $\occStatek$ at time $k$ is considered either occupied or free: $\occStatek \in \{ \occEvent, \freeEvent \}$. The grid map updates the grid cell states when a new measurement arrives.
For this purpose, an inverse sensor model assigns a discrete, binary occupancy probability $p_{\zp}(\occStatep|\zp)$ individually to each grid cell based on the measurement $\zp$ at time $k+1$. The result is called a measurement grid.
To give a practical example, consider a laser range measurement consisting of several laser beams and the resulting measurement grid as depicted in Fig. \ref{fig:measGrid}. 
The inverse sensor model can be a heuristic model or the result of a machine learning process \cite{Thrun2005}. 
Usually, the position of the robot or vehicle in the grid map is estimated by a dead reckoning approach \cite{Dietmayer2015}. 

The posterior occupancy probability $\postop(\occStatep)$ at time $\kp$ results from the last posterior occupancy probability $\postok(\occStatek)$ at time $k$ and the measurement-based estimate $p_{\zp}(\occStatep|\zp)$ through \cite{Thrun2005}

\begin{align}
&\postop(\occStatep) = \nonumber \\
&\frac{p_{\zp}(\occStatep|\zp) \cdot \postok(\occStatek)}
{p_{\zp}(\occStatep|\zp) \cdot \postok(\occStatek)  +
   p_{\zp}(\noccStatep|\zp) \cdot \postok(\noccStatek)  },
\label{eq:BBF}
\end{align}
where $p(\noccState) = 1- p(\occState)$ denotes the probability of the counter event of occupied or free, respectively.

This is often referred to as binary Bayes filter due to the binary nature of the estimated state. Equation \eqref{eq:BBF} holds if the prior probability for occupancy and free is equal, the measurements are independent of each other and the grid cell state does not change over time. 
An alternative approach is the forward sensor model, which estimates for each grid cell the occupancy likelihood function $\likep(\zp|\occStatep)$ for the two feasible occupancy events $\occStatep \in \{ \occEvent, \freeEvent \}$.
Then the update under the same assumptions as for \eqref{eq:BBF} is given by
\begin{align}
&\postop(\occStatep) = \nonumber \\
&\frac{ \likep(\zp|\occStatep) \cdot \postok(\occStatek)}
{\likep(\zp|\occStatep) \cdot \postok(\occStatek)
 +  \likep(\zp|\noccStatep) \cdot \postok(\noccStatek) }.
\label{eq:BBF_forward}
\end{align}

%The occupancy likelihood is a density functions in regard to the measurement $\zp$; their physical unit is a density in the measurement space. 
Modeling likelihoods is more complicated than designing inverse sensor models and usually also computationally more expensive. 

Equations \eqref{eq:BBF} and \eqref{eq:BBF_forward} can be generalized to 
\begin{align}
\postop(\occStatep) =
\frac{ \alpha_{\zp} \cdot \postok(\occStatek)}
{ \alpha_{\zp} \cdot \postok(\occStatek)
 + \postok(\noccStatek)  }, 
\label{eq:BBF_gen}
\end{align}
where $\alpha_{\zp}$ is the single measurement based occupancy probability ratio 
\begin{align}
\alpha_{\zp} = \frac{p_{\zp}(\occStatep|\zp)}{p_{\zp}(\noccStatep|\zp)}, \,\,\, 
p_{\zp}(\noccStatep|\zp) > 0, 
%\label{eq:alpharatio_inv} % Never referenced, so I commented this label which is defined twice.
\end{align}
or the likelihood ratio
\begin{align}
\alpha_{\zp} = \frac{\likep(\zp|\occStatep)}{\likep(\zp|\noccStatep)}, \,\,\,
\likep(\zp|\noccStatep) >0,
%\label{eq:alpharatio_inv} % Never referenced, so I commented this label which is defined twice.
\end{align}
respectively. As a conclusion, the binary Bayes filter requires either a likelihood ratio or a probability ratio for the update step. It will be shown later in Sect. \ref{sec:proof} that the binary Bayes filter \eqref{eq:BBF_gen} is a special case of the presented PHD/MIB filter, namely for the assumption of zero velocity in a deterministic process model. 

\subsection{Dynamic Grid Mapping}
Since the assumption of a stationary environment is not realistic for typical traffic scenarios, several approaches to
integrate object movement into grid maps have been proposed recently. This section compares four contributions from Danescu et al. \cite{Danescu2011}, Tanzmeister et al. \cite{Tanzmeister2014}, N\`{e}gre et al. \cite{Negre2014} and Nuss et al. \cite{Nuss2015}.  

\begin{figure}[t]
\begin{tabular}{cc}
  \includegraphics[width=0.45\columnwidth]{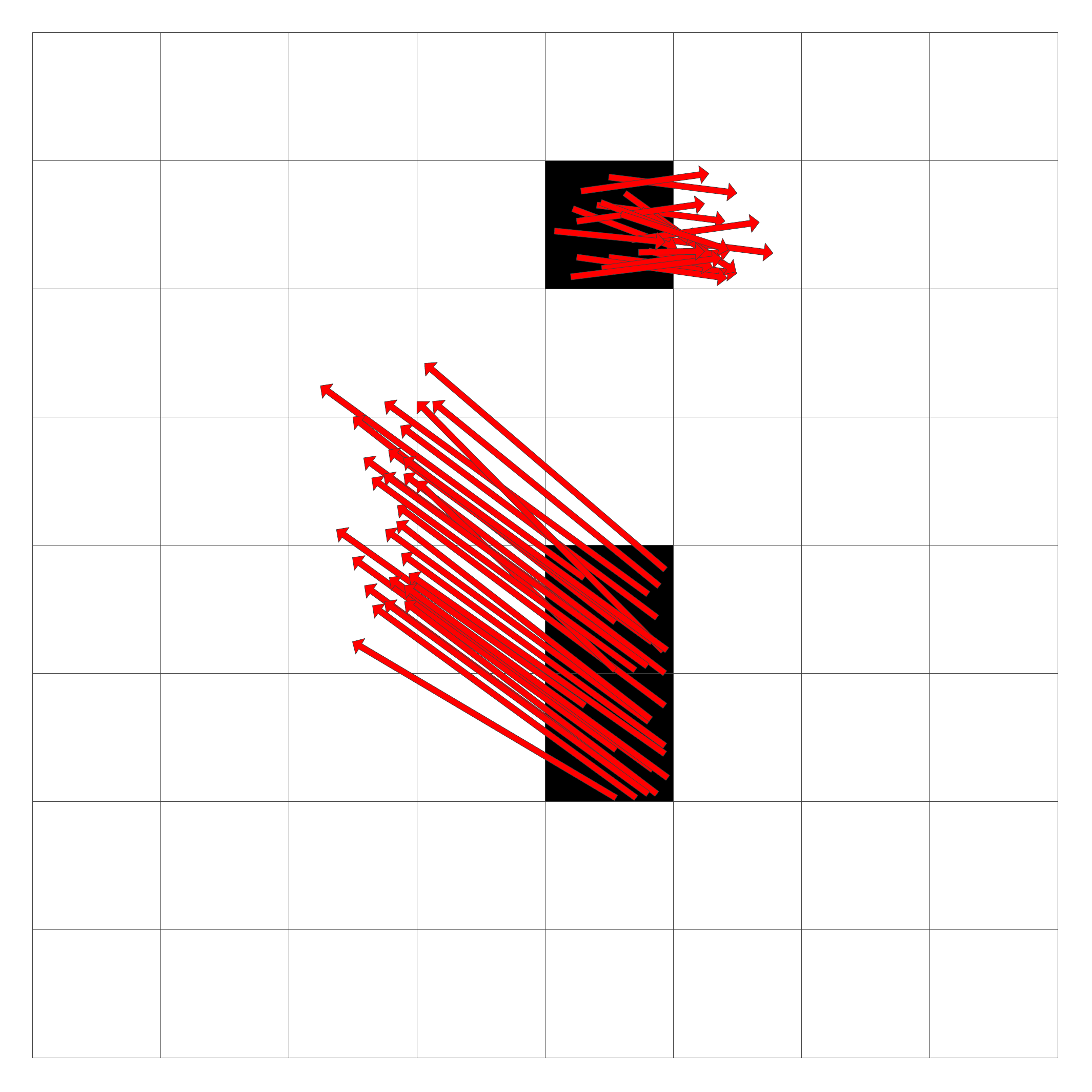} &   \includegraphics[width=0.45\columnwidth]{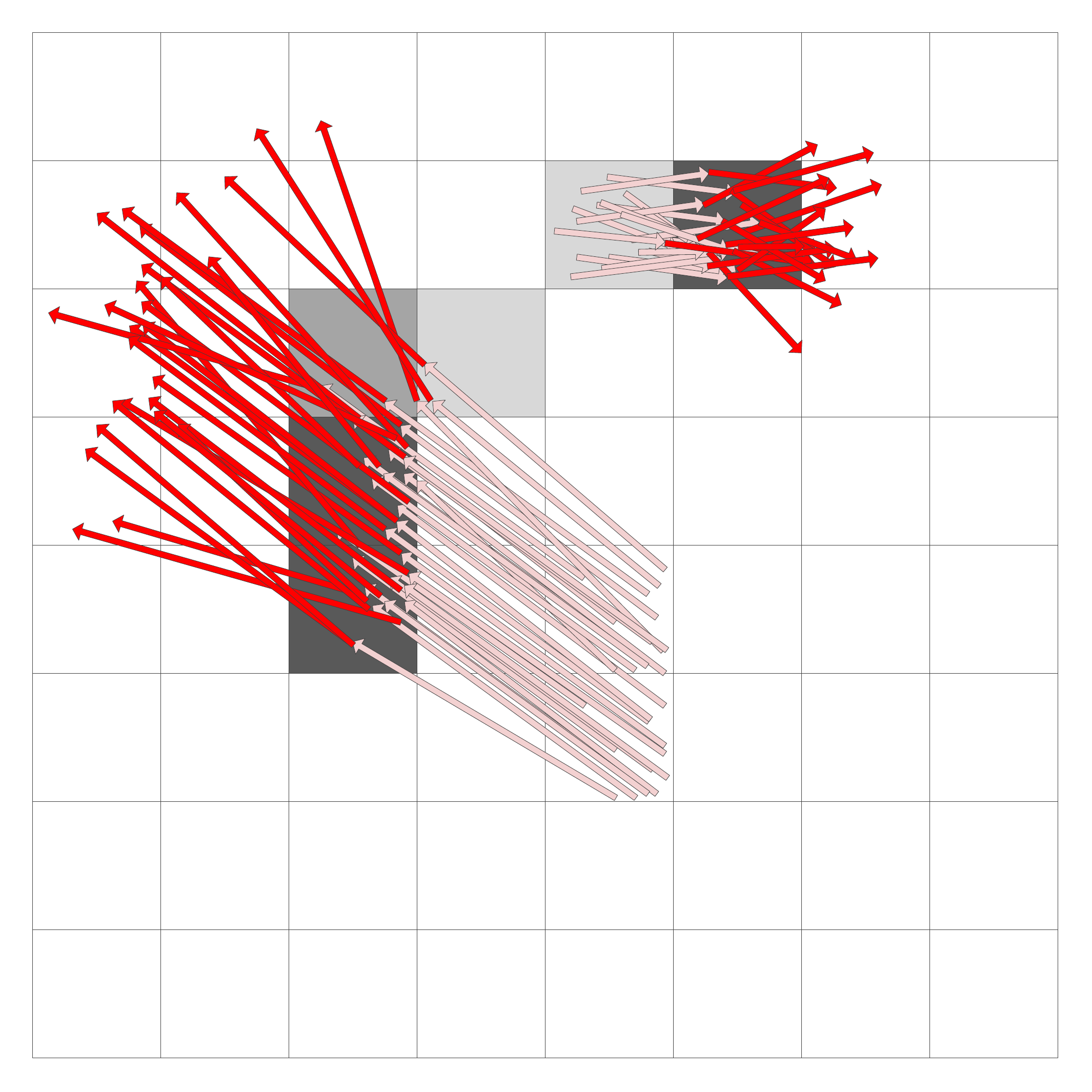} \\
  {\footnotesize (a) Posterior at $k$} & {\footnotesize (b) Prediction for $k+1$}\\[6pt]
  \includegraphics[width=0.45\columnwidth]{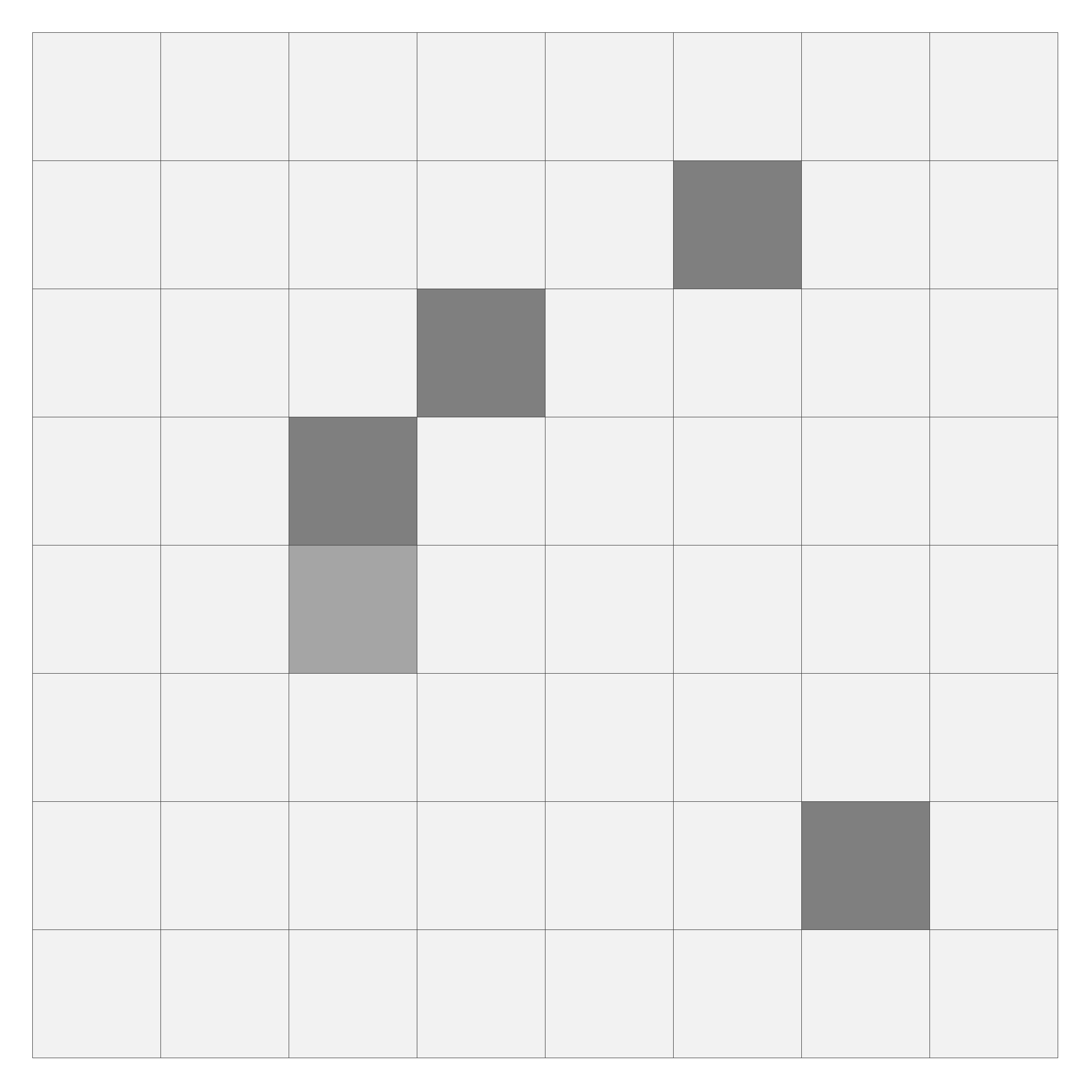} &   \includegraphics[width=0.45\columnwidth]{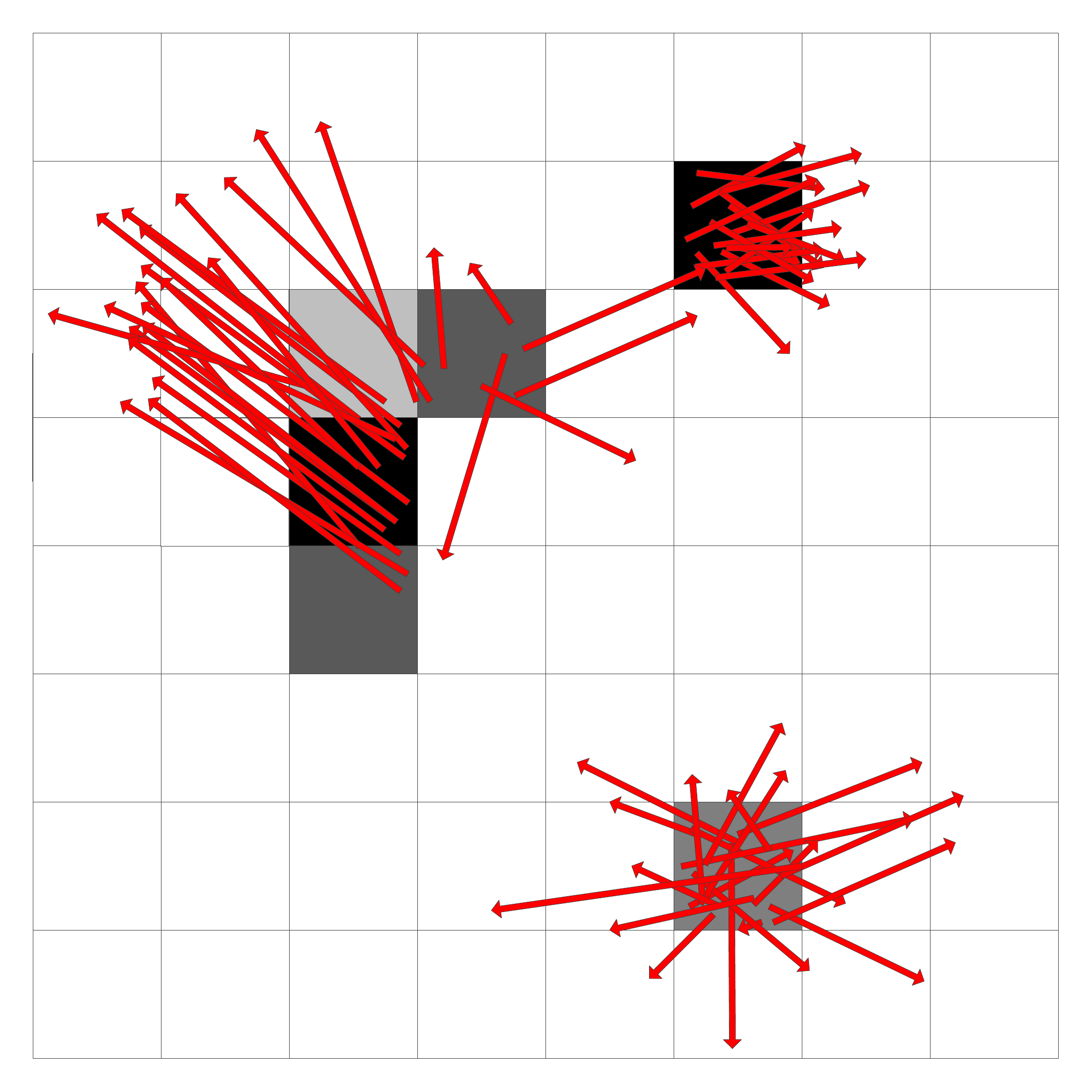} \\
  {\footnotesize (c) Measurement grid at $k+1$} & {\footnotesize (d) Posterior at $k+1$}\\[6pt]
\end{tabular}
\caption{Different states of dynamic grid map estimation recursion.}
\label{fig:particles}
\end{figure}

All mentioned publications about particle-based dynamic grid maps estimate the occupancy probability and the dynamic state of grid cells in the vehicle environment. Further, all publications describe an algorithm consisting of a prediction and an update step as depicted in Fig. \ref{fig:particles} and apply a resampling step to avoid degeneration. 

\subsubsection{State Representation}
All mentioned publications represent the dynamic state of grid cells with particles, but the interpretation of a particle differs: \cite{Danescu2011} and \cite{Nuss2015} directly use the number of particles or the sum of particle weights in a grid cell as a measure for the occupancy probability of the grid cell. In contrast, \cite{Negre2014} propagates an additional discrete probability distribution for the events free, static occupancy and dynamic occupancy for each grid cell. The particles then represent a velocity distribution for the dynamic case. The same events are used in \cite{Tanzmeister2014} within a Dempster-Shafer framework \cite{Dempster1968}.
To avoid aliasing problems, particles represent velocity and position of an occupancy in a grid cell in all mentioned publications, so the dynamic state of a grid cell is four-dimensional.

\subsubsection{Prediction Step}
All mentioned publications assume a process model with constant velocity and constant direction and propagate each single particle accordingly. All particles that are predicted into a certain grid cell represent the predicted dynamic state of the grid cell.
However, the exact quantitative reasoning about the resulting predicted occupancy probability varies. Intuitively, the higher the number of particles or particle weights predicted into a grid cell, the higher is the predicted occupancy probability. An example is depicted in Fig. \ref{fig:particles}b.

\subsubsection{Update Step}
Updating the occupancy probability of a grid cell with a measurement grid is generally a binary Bayes problem and solved either by equation \eqref{eq:BBF} or \eqref{eq:BBF_forward} or by equivalent update steps in the Dempster-Shafer framework \cite{Tanzmeister2014}, \cite{Nuss2015}. Due to the lack of a mathematically rigorous definition of a particle, all publications use different methods to normalize the particle weights in a grid cell after the update step to provide a consistent representation of the occupancy and the dynamic state of a grid map. 

\subsubsection{Resampling}
All mentioned publications apply a resampling step to avoid degeneration. Similar to classic particle filters, the resampling step chooses to eliminate some particles and reproduce others instead, based on their weight. After the resampling step, all particles are assigned the same weight.

\subsubsection{Initialization}
If a measurement grid cell provides a high occupancy probability (or occupancy likelihood, in the forward case), but no particles were predicted into the corresponding grid cell, new particles must be initialized to represent the dynamic state of the grid cell. The initial distribution depends on the environment setting, but usually the velocity of movements is limited, e.g. by the maximum speed of a vehicle.
  
Neither \cite{Danescu2011} nor \cite{Tanzmeister2014} describe the initialization step to any further detail than mentioned here. In a realistic scenario, a grid cell is not either empty or fully populated, but mostly something in between. Then the question arises how to divide the weight between predicted and initialized particles. Intuitively, the weight for newly initialized particles should rise with increasing measurement occupancy and decreasing predicted occupancy. Heuristic examples are provided by \cite{Negre2014} and \cite{Nuss2015}.

\subsubsection{Occluded Areas}
In practical applications, a grid map contains a high ratio of occluded and therefore unobserved grid cells. Populating unknown areas of the grid map with particles would result in a huge computational load. To avoid this, all mentioned publications only initialize particles in grid cells with a certain measured occupancy probability.

\subsection{Discussion}
The discussed particle-based BOFs show promising results. However, from a theoretical point of view many open questions remain.
A prerequisite for Bayesian
state estimation is the definition of a state space, a stochastic process describing the state transition and a
stochastic observation process. All mentioned papers directly describe the propagation of particles without 
defining the estimation problem first. As a result, it is unclear what a particle represents. All mentioned publications explain that a particle represents a hypothesis for the dynamic state of an individual grid cell. However, during the prediction step, particles from various cells are predicted
into another grid cell and jointly represent the state of the destination cell. The particles are not assigned to a specific object, instead they represent a hypothesis for the existence and state of a whole group of objects. 

In other words,
the particles jointly represent a set of occupied grid cells, where the number of occupied grid cells is a random process
itself and must be estimated too. This cannot be explained with single-object Bayesian estimation theory. As a
consequence, previous work cannot motivate prediction or update equations for a
well-defined estimation problem. Especially the initialization of new particles remains unclear.

From a theoretical point of view, an environment containing a random but limited number of objects is a random finite set (RFS) \cite{Mahler2007}. The finite set statistics (FISST) are a mathematical framework providing a basis for
Bayesian state estimation of multiple objects. The following section gives an introduction to the basics of FISST
required for the derivation of dynamic grid mapping as an RFS estimation problem.

\section{Random Finite Set Statistics}
\label{sec:rfs}
This section outlines the main concepts of finite set statistics and the multi-object Bayes
filter. For further details, the reader is referred to \cite{Mahler2007} or \cite{Ristic2013c}.

A random finite set (RFS) is a finite set-valued random variable, i.e., a realization of
an RFS consists of a random number of points or objects whose individual states are
given by random vectors $\x \in \stateSpace$ where $\stateSpace$ denotes the single-object
state space. Thus, an RFS is represented by
\begin{align}
\sX = \{\x^{(1)}, \hdots, \x^{(n)} \} \nonumber
\end{align}
where $n\geq0$ is a random variable and the special case $n=0$ results in the empty set
$\sX=\emptyset$.

The cardinality distribution of an RFS is given by an arbitrary discrete distribution
and the probability for an RFS representing exactly $n$ objects is denoted by $\rho(n)$.
For each cardinality $n>0$, the RFS contains a set of probability density functions (PDFs)
\begin{align}
\{f_n(\x^{(1)}, \hdots, \x^{(n)}), \, n \in \mathbb{N} \,\, |\,\, \rho(n) > 0 \} \nonumber,
\end{align}
i.e., the RFS supports several different state distributions for a single cardinality.
Since an RFS is order independent, the multi-object probability density function (MPDF) is given
by
\begin{align}
\label{eq:mpdf}
&\pi(\sX = \{\x^{(1)}, \hdots, \x^{(n)} \}) = \notag \\
&\quad \quad \quad \quad \begin{cases}
    \rho(0)             & \text{if } \sX = \emptyset,  \\
     n! \cdot \rho(n) \cdot f_n(\x^{(1)}, \hdots, \x^{(n)})  & \text{otherwise},
\end{cases}
\end{align}
where the factor $n!$ accounts for all possible permutations of the vectors $\x^{(1)}, \hdots, \x^{(n)}$.

Since the number of objects is also a random variable, the set integral \cite{Mahler2007}
\begin{align}
&\int \spi(\sX) \delta \sX =  \spi(\emptyset) +  \notag \\
&\quad \quad \quad \quad \sum_{n=1}^{\infty} \frac{1}{n!} \int  \spi(\x^{(1)}, \hdots, \x^{(n)}) d\x^{(1)} \cdots d\x^{(n)}
\label{eq:setIntegral}
\end{align}
has to be applied for the integration over an MPDF.

\subsection{Multi-Object Bayes Filter}
Conventional multi-object tracking is typically realized using several
instances of a Kalman filter \cite{Kalman1960a}. This provides an analytical solution to the
single-object Bayes filter in case of Gaussian distributed states and measurements as well
as linear motion and measurement models. The multi-object Bayes filter \cite{Mahler2007}
is a generalization of the single-object Bayes filter which handles the uncertainty in
the number of objects in a mathematically rigorous way.

If the multi-object density at time $k$ is given by $\spikraw(\sXk)$, the predicted multi-object density
is obtained by applying the Chapman-Kolmogorov equation:
\begin{align}
\spipredraw (\sXp) = \int \trans(\sXp|\sXk)\spikraw(\sXk) \delta \sXk .
\end{align}
Here, $\trans(\sXp|\sXk)$ denotes the multi-object transitional density which captures
the appearance and disappearance of objects in addition to the movement of persisting objects.
For a shorter notation, the index "$+$" expresses a prediction step from time $k$ to time $k+1$, often noted as  $k+1|k$. 

The measurement update of the predicted multi-object density using a set of measurements $\sZp$
is realized by applying Bayes' rule to yield
\begin{align}
\spipraw(\sXp|\sZp) \! = \! \frac{ \likemultiraw(\sZp|\sXp) \spipredraw(\sXp) }{ \int \likemultiraw(\sZp|\sXp) \spipredraw(\sXp) \delta \sXp},
\label{eq:multi_obj_update}
\end{align}
where the integral in the denominator is a set integral as defined in Eq. \eqref{eq:setIntegral}. Similar to the
multi-object transitional density in the prediction step, the multi-object likelihood function $\likemultiraw(\sZp|\sXp)$
has to incorporate the uncertainty of the measurement process, i.e., it has to model missed detections
and false alarms. 

A realization of the multi-object Bayes filter is possible using Sequential Monte Carlo (SMC) methods
(e.g. \cite{Vo2005, Mahler2007, Reuter2013b}) or Generalized Labeled Multi-Bernoulli distributions \cite{Vo2013, Vo2014a}.
Further, several approximations like the Probability Hypothesis Density (PHD) filter \cite{Mahler2003,Vo2006c,Ristic2013c}, the
Cardinalized PHD filter \cite{Mahler2007a,Vo2007}, the Cardinality Balanced Multi-Bernoulli Filter \cite{Vo2009}
and the Labeled Multi-Bernoulli filter \cite{Reuter2014} have been proposed during the last decade.

\subsection{PHD and Bernoulli RFS}
The PHD filter approximates the full multi-object density using the first statistical moment which is
given by its intensity distribution or probability hypothesis density (PHD) \cite{Mahler2003}:
\begin{align}
&\D(\x) = E\left\{\sum_{{w} \in \sX} \delta(\x -{w}) \right\} = \int \sum_{{w} \in \sX} \delta(\x -{w}) \spi(\sX) \delta \sX .
\end{align}
Here, $E\{\cdot\}$ denotes the expectation.
Since $\D(\x)$ is an intensity distribution, the integral over $\D(\x)$ corresponds to the expected
number of targets in this area.

An important multi-object distribution for the remainder of this contribution is the Bernoulli RFS.
A Bernoulli RFS \cite{Mahler2007} is typically used to model scenarios where an object either exists with an existence
probability $r$ or does not exist with a probability of $1-r$. If the object exists, its spatial
distribution is given by the single-object PDF $p(\x)$. Consequently, the multi-object probability density
follows
\begin{align}
\spi(\sX) = 
\begin{cases}
     1 -r      & \text{if } \sX = \emptyset,  \\
     r \cdot p(\x)   & \text{if } \sX = \{ \x \},\\
     0   & \text{if } |\sX| \geq 2.
\end{cases} \nonumber
\label{eq:bernoulli_rfs}
\end{align}
The intensity function or PHD of a Bernoulli RFS, which corresponds to the first statistical moment,
is given by the product of the existence probability and the spatial distribution \cite{Mahler2007}:
\begin{align}
D(\x) = r \cdot p(\x).
\end{align}

\section{The Probability Hypothesis Density / Multi-Instance Bernoulli (PHD/MIB) Filter}
\label{sec:phdmib}

\begin{figure}[t]
  \centering
  \includegraphics[width=\columnwidth]{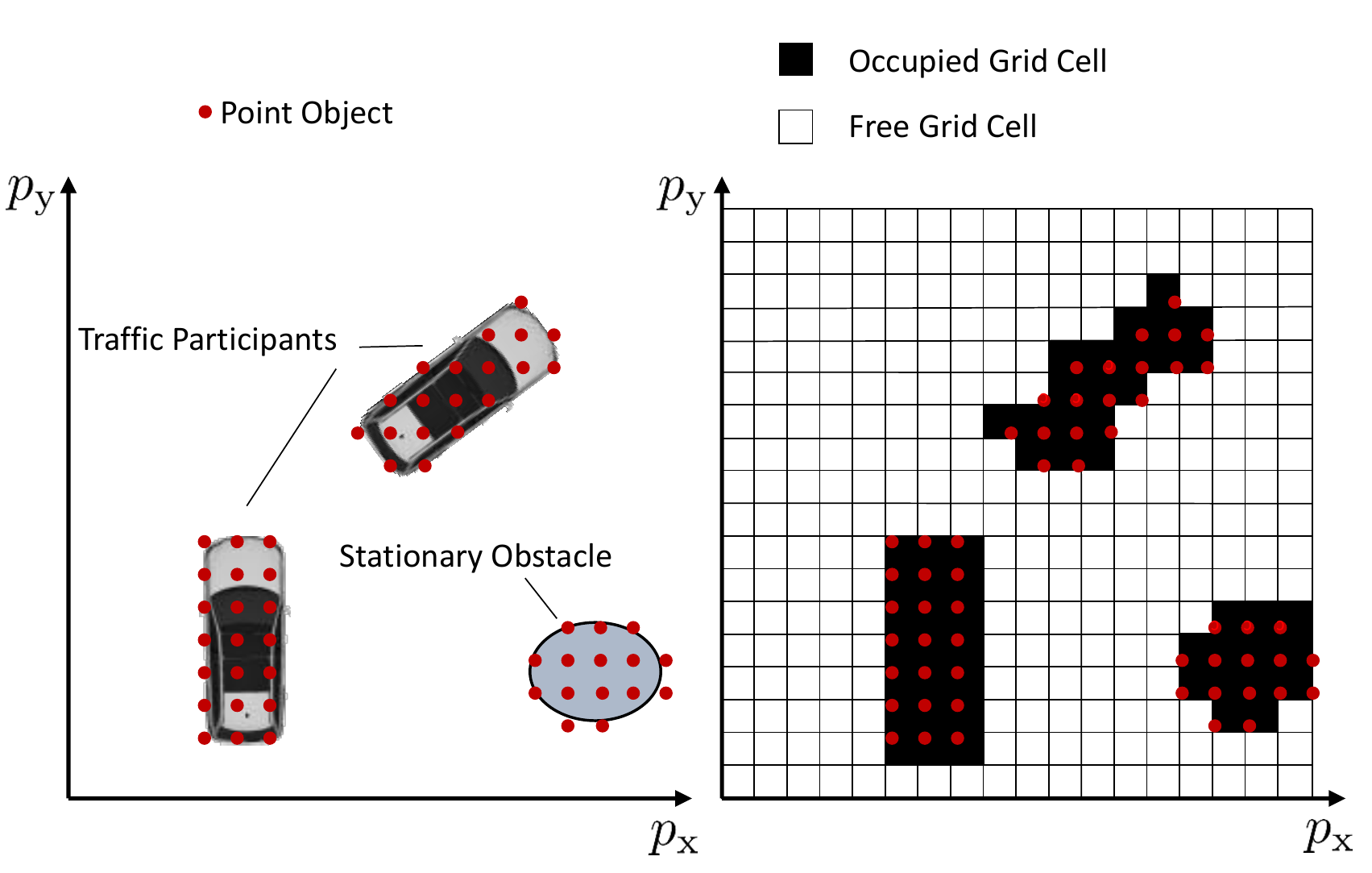}
	\caption{Left: Real-world objects and corresponding point objects. Right: The number of point objects per real-world object equals the number of grid cells occupied by the real-world object.}
	\label{fig:point_object}
\end{figure}

This section proposes to model the dynamic grid map using random finite set theory which facilitates to combine
the Bernoulli filter and the PHD filter to recursively estimate the state of dynamic grid cells. These filters cover fundamental problems like object initialization or modeling of heterogeneous measurements in an elegant way. The result is a recursion called probability hypothesis density / multi-instance Bernoulli (PHD/MIB) filter. First, the environment model and the estimation problem are defined. Based hereon, the filter steps are outlined and the prediction and update equations of the PHD/MIB filter are derived.

\subsection{Environment Definition and Filter Recursion Outline}
The PHD/MIB filter estimates a random finite set consisting of so-called point objects. The relation between point objects and real-world objects depends on an underlying grid map and is shown in Fig. \ref{fig:point_object}. A real-world object consists of at least one but possibly several point objects. The number of corresponding point objects per real-world object equals the number of grid cells occupied by the real-world object.

To provide an example for the state $\x$ of a point object, let $x$ be a two-dimensional position and a two-dimensional velocity:
\begin{align}
\x = [p_\text{x} \, p_\text{y} \, v_\text{x} \, v_\text{y}]\transp.
\end{align}
However, the object state can be arbitrarily extended and can include additional attributes such as object height, color, semantic class, etc.

\begin{figure}[t]
  \centering
  \includegraphics[width=\columnwidth]{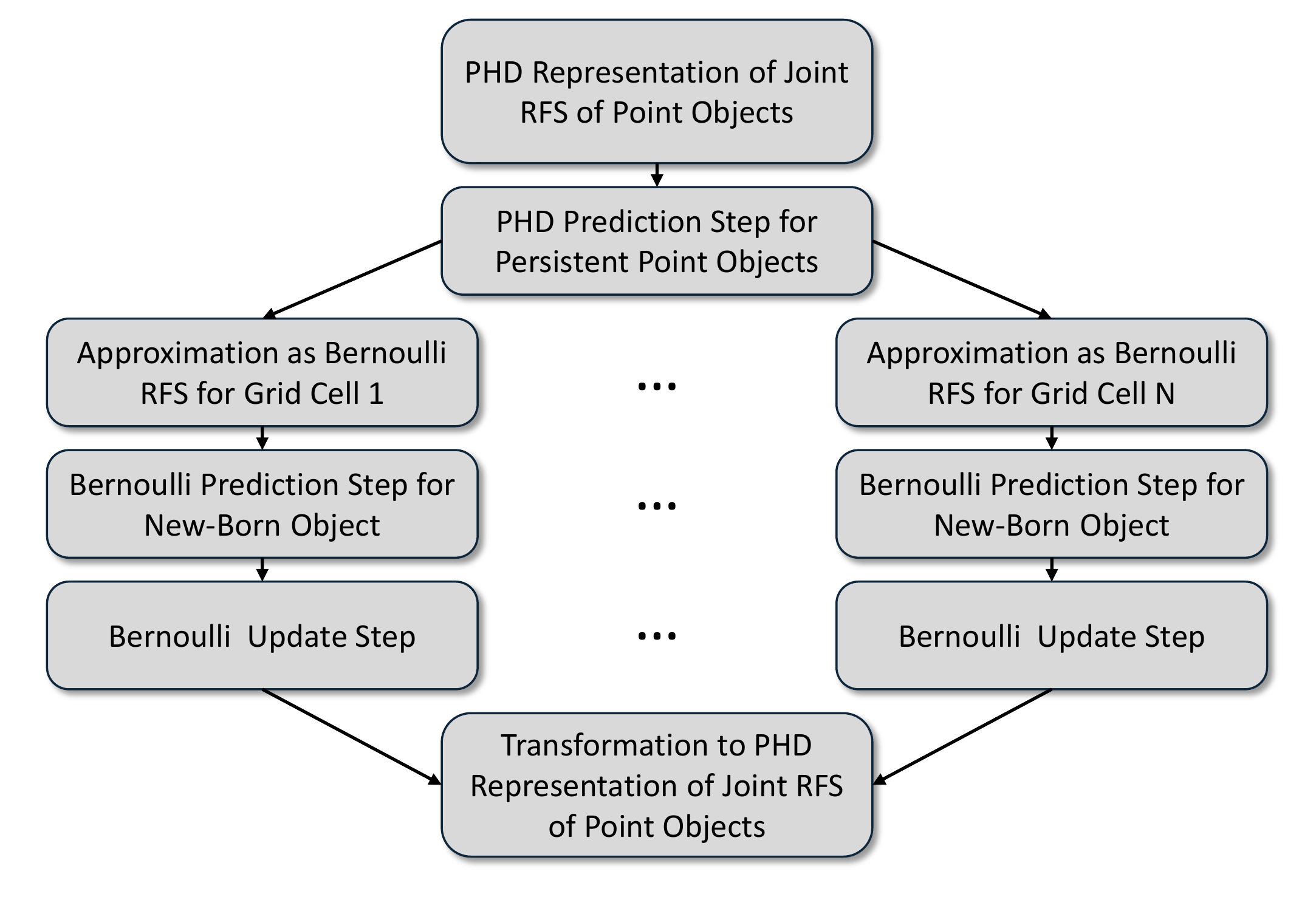}
	\caption{ Processing scheme of the PHD/MIB filter.}
	\label{fig:FilterOverview}
\end{figure}

The goal of the PHD/MIB filter is to estimate the multi-object state of the point objects in a vehicle environment, which includes an estimate of the occupancy state of grid cells, as will be explained below. In the course of the filter recursion, the PID/MIB filter represents the random finite set of point objects in different forms. An overview of the filter recursion is depicted in Fig. \ref{fig:FilterOverview}. The posterior state is uniquely represented by its probability hypothesis density (PHD). The PHD/MIB filter prediction step simply applies the PHD filter prediction. In order to update the predicted RFS state with a measurement grid, the PHD/MIB filter approximates the point object state in each grid cell as a Bernoulli RFS and carries out the update step independently for each grid cell.
Finally, the PHD/MIB filter transfers all instances of Bernoulli sets to a joint PHD to represent the posterior state. The individual filter steps are detailed below.

\subsection{Multi-Object State Transition in PHD representation}
\begin{figure}[t]
  \centering
  \includegraphics[width=\columnwidth]{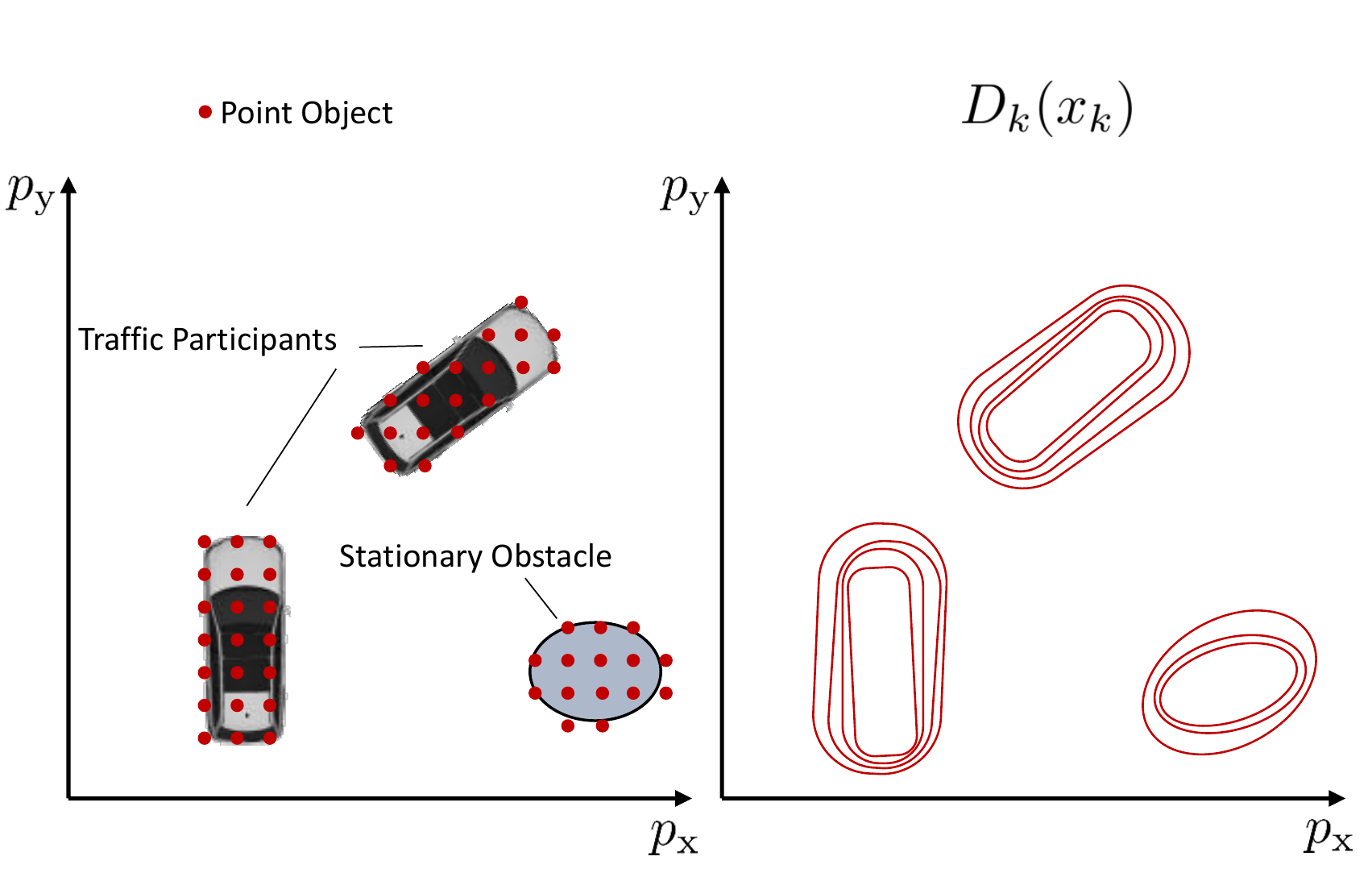}
	\caption{ Exemplary PHD of a traffic situation. Here, only the subspace of the two-dimensional position is visualized.}
	\label{fig:PHDrep}
\end{figure}

The PHD/MIB filter expresses the posterior multi-object state of time step $k$ by its PHD $\Dk(\xk)$.
%\begin{align}
%\Dk(\xk) : \mathbb{R}^4  \rightarrow \mathbb{R}.
%\end{align}
%
In practical applications, PHDs are commonly represented by particles or Gaussian mixtures. However, the following derivation is independent of its practical representation form.
Figure \ref{fig:PHDrep} shows an exemplary PHD of a traffic situation using contour lines of a Gaussian mixture.
The birth process of point objects is defined by the birth PHD $\gamma_\text{b}(\xp)$ and the persistence probability of each point object is denoted by $\pS$.
The standard prediction step of a PHD filter is then given by \cite{Mahler2003}
\begin{align}
\Dpred(\xp) = \,\, &\gamma_\text{b}(\xp) + \notag \\
&\pS  \int \trans(\xp|\xk)\Dk(\xk) d \xk,
\end{align} 
where $\trans(\xp|\xk)$ is the single-object transition density. An example for a process model which defines a transition density is the constant velocity (CV) model given by
\begin{align}
\label{eq:process}
\xp = 
\begin{pmatrix}
1 & 0 & T & 0 \\
0 & 1 & 0 & T \\
0 & 0 & 1 & 0 \\
0 & 0 & 0 & 1 \\
\end{pmatrix}
\xk + \xi_{k}, 
\end{align} 
where $\xi_{k}$ is the process noise and $T$ is the time interval between $k$ and $k+1$.

Since new-born objects will be handled in Bernoulli form, the prediction step of the PHD/MIB filter is only required to handle
persisting objects and consequently simplifies to
\begin{align}
\Dpredp(\xp) = \pS  \int \trans(\xp|\xk)\Dk(\xk) d \xk.
\label{eq:prediction_persistent}
\end{align}
Here, the lower index p symbolizes the affiliation to persistent objects.

\subsection{Approximation as Bernoulli Distributions}
The PHD/MIB filter approximates the predicted PHD $\Dpredp(\xp)$ of persisting objects by multiple instances
of independent Bernoulli RFSs, one for each grid cell. The interpretation is that each grid cell can either be occupied (a point object exists in the grid cell) of free (no point object exists in the grid cell). Each Bernoulli RFS instance models the possibility of object birth and the observation process including clutter. The birth process model is identical to the standard Bernoulli filter. The clutter process is different and will be derived for the special form of a measurement grid. The following part describes the Bernoulli filter steps for a single grid cell denoted by $c$. 

The predicted Bernoulli RFS for a persistent point object in grid cell $c$ is given by
\begin{align}
\spipredp (\sXp) \!=\! 
\begin{cases}
\experspredn        & \!\!\!\! \text{if } \sXp \!=\! \emptyset,  \\[.5em]
\experspred \cdot \pdfperspred(\xp)  &\!\!\!\! \text{if } \sXp \!= \!\{ \xp \},
\end{cases}
\end{align}
where $\experspred$ is the predicted existence probability of a persistent point object, $\pdfperspred(\xp)$ is the predicted PDF of its state, and $\experspredn = 1-\experspred$. The upper index $\ci$ denotes values which are different for individual cells. From the definition of the PHD it follows that
\begin{align}
\experspred = \text{min} \left( \int\limits_{\xp \in c}\Dpredp(\xp) d\xp ,\ 1 \right),
\label{eq:predExistenceProb}
\end{align}
where the set $\{ \xp | \xp \in c \}$ is the subset of $\mathbb{R}^4 $ associated to grid cell $c$.
Here, the limitation to a maximum value of $1$ is a required approximation since the PHD prediction does not consider
that each grid cell cannot be occupied by more than one point object. Further, the spatial distribution within the grid
cell is given by
\begin{align}
\pdfperspred(\xp) = \frac{\Dpredp(\xp)}{\experspred}\text{ for all }\xp \in c 
\,\,\, \text{and} \,\,\,  \experspred>0. 
\label{eq:predDistribution}
\end{align}

\subsection{Bernoulli RFS Birth Model}
The standard Bernoulli filter defines the following birth model \cite{Ristic2013c}: If no object exists at time step $k$, the existence probability of a new-born object at time step $k+1$ is given by the prior birth probability $p_\text{B}$. If an object exists at time step $k$, the existence probability of a new-born object at time $k+1$ is zero, independent of the survival of the old object. In the case of a new-born object, the single-object spatial birth PDF is given by $\pb(\xp)$.

The PHD/MIB filter assumes the same birth process as the standard Bernoulli filter. This results in the predicted existence probability $\exbornpred$ of a new-born object with
\footnote{Formally, the probability of a birth event in time step $k+1$ depends on the existence probability of an object in time step $k$ for the Bernoulli filter \cite{Ristic2013c}. For simplicity, equation \eqref{eq:predicted_ex_born} neglects that fact and directly relates to the predicted existence probability for time step $k+1$. This is appropriate for process models with a high persistence probability ($\pS \approx 1$).} 
\begin{align}
\exbornpred =  p_\text{B} \cdot \left(1- \experspred \right).
\label{eq:predicted_ex_born}
\end{align}
%

%\begin{align}
%\exbornpred =  p_\text{B} \cdot \left(1- \frac{\experspred}{\pS}\right).
%\label{eq:predicted_ex_born}
%\end{align}
%
Considering both cases of a persistent or new-born object leads to the predicted Bernoulli RFS
\begin{align}
&\spipred (\sXp) =  \nonumber \\ 
&\begin{cases}
1 - \exbornpred - \experspred          &\text{if } \sXp = \emptyset,  \\
\exbornpred \cdot \pb(\xp) + \experspred \cdot \pdfperspred(\xp)   &\text{if } \sXp = \{ \xp \}.
\end{cases}
\label{eq:predicted_bern}
\end{align}
Note that the predicted Bernoulli RFS $\spipred (\sXp)$ represents both the predicted dynamic distribution and the predicted occupancy probability of the corresponding grid cell. The predicted occupancy probability $\predok^\ci(\occEvent_{\kp})$ equals the combined predicted existence probability $\expred$ of a persistent or a new-born point object in the grid cell, so by definition of the Bernoulli RFS it follows:
\begin{align}
\predok^\ci(\occEvent_{\kp}) = \expred  =  \exbornpred + \experspred.
\end{align}

\subsection{Bernoulli Observation Process}
The standard Bernoulli filter expects Poisson distributed clutter detections \cite{Ristic2013c}. Since in practical applications a measurement grid cell usually does not contain more than one measurement, this is not feasible.

The PHD/MIB defines the following Bernoulli RFS observation process instead: A measurement grid map provides an observation for each grid cell based on sensor data at time step $k+1$. The observation process of one grid cell is assumed independent of other grid cells.

In each grid cell either one measurement $\zpcell$ or no measurement occurs. The probability that a measurement occurs in the occupied grid cell $c$ is the cell-specific and time-dependent true positive probability $\cellTP \in (0,1)$. The probability that a measurement occurs in the empty cell $c$ is the false positive probability $\cellFP \in (0,1)$. The PDF of a false positive measurement is given by the clutter density $\pc(z)$. 

A true positive measurement is associated to the point object in the grid cell with the association probability $\pA$. In this case its distribution is defined by the single-object likelihood function $\likecell$. If the measurement is not associated to the point object, its PDF is also defined by the clutter density $\pc(z)$.

In practical terms, each measurement grid cell $c$ contains the following data: the individual, time-dependent true positive and false positive probabilities $\cellTP$ and $\cellFP$. In case a measurement occurred in the cell it additionally provides the single-object likelihood function $\likecell$ and the corresponding association probability $\pA$.

The resulting multi-object likelihood $\likemulti(\sZp|\sXp)$ for the measurement grid cell $c$ calculates to
\begin{align}
&\likemulti(\sZp= \emptyset | \sXp = \emptyset ) = \cellFPn,	\nonumber \\
&\likemulti(\sZp=\{\zp\} | \sXp = \emptyset ) = \cellFP \cdot \pc(\zp),  	\nonumber \\
&\likemulti(\sZp= \emptyset |\sXp = \{ \xp \}) = \cellTPn, \nonumber
\end{align}
% special break to for IEEE trans style
\begin{align}
&\likemulti(\sZp=\{\zp\}  |\sXp = \{ \xp \}) = \nonumber\\
& \,\,\,\,\,\, \cellTP \underbrace{\left[ \pA \cdot  \likecell + \pAn \cdot \pc(\zp) \right]}_{=: \likecellA} .
\label{eq:multi_obj_likelihood}
\end{align}

\subsection{Multi-Object Bayes Update Step}
Ultimately, inserting the predicted Bernoulli RFS \eqref{eq:predicted_bern} and the multi-object likelihood \eqref{eq:multi_obj_likelihood} into the general multi-object update \eqref{eq:multi_obj_update} leads to the posterior multi-object PDF $\spip(\sXp|\sZp)$:
\begin{align}
&\spip  (\sXp = \emptyset|\sZp = \emptyset) = \frac{\cellFPn \cdot \overline{r}^\ci_{\kpred}}{\normempty},   \nonumber\\
&\spip  (\sXp = \{ \xp \}|\sZp = \emptyset) = \frac{\cellTPn \cdot \spipred (\{ \xp \} )}{\normempty},  \nonumber\\
&\spip  (\sXp = \emptyset |\sZp = \{ \zp\}) = \frac{\cellFP \cdot \pc(\zp) \cdot \overline{r}^\ci_{\kpred}}{\normz}, \nonumber\\
&\spip  (\sXp = \{ \xp \} |\sZp = \{ \zp\}) = \nonumber\\
& \,\,\,\,\,\,\,\,\,\,\,\, \frac{\cellTP \cdot  \likecellA \cdot \spipred( \{\xp\} )}{\normz}.
\label{eq:filter_update}
\end{align}

Using \eqref{eq:setIntegral}, the normalization constants calculate to 
\begin{flalign}
\normempty 
&=  \int \likemulti(\sZp = \emptyset|\sXp) \spipred(\sXp) \delta \sXp  \nonumber \\
&= \cellFPn \cdot \overline{r}^\ci_{\kpred} + \cellTPn  \cdot {r}^\ci_{\kpred} 
\label{eq:normalization_empty}
\end{flalign}
and
\begin{flalign}
\normz &=  \int \likemulti(\sZp = \{ \zp\}|\sXp) \spipred(\sXp) \delta \sXp  \nonumber \\
&= \cellFP \cdot \pc(\zp) \cdot \overline{r}^\ci_{\kpred}  \nonumber\\ 
&  + \cellTP \! \int \! \likecellA  \cdot \spipred( \{\xp\}) d\xp. 
\label{eq:normalization_meas}
\end{flalign}

The posterior $\spip(\sXp)$ is a Bernoulli RFS and represents both the dynamic state and the occupancy probability of the corresponding grid cell. According to the definition of a Bernoulli RFS, the posterior occupancy probability of grid cell $c$ is 
\begin{align}
\postop^\ci(\occEvent_{\kp}) = \exkp = \int\limits_{\xp \in c} \spip(\{ \xp \}) \,d\xp.
\label{eq:occprob_integral}
\end{align}

After the update step, the PHD/MIB filter transforms all Bernoulli RFS instances into a joint PHD again. The joint PHD is simply the sum of all Bernoulli RFS instances:

%\textbf{SR: Mathematically not correct, PHD cannot be the sum of Bernoulli RFSs, just replace big X by small x to only include singletons?}

%\textbf{DN: FIXED, to be verified}

\begin{align}
\Dp(\xp) = \sum_{c = 1}^C \spip(\sXp=\{ \xp \}), 
\label{eq:jointPDH}
\end{align}
where $c$ denotes the index of the corresponding grid cell of each Bernoulli RFS instance and $C$ is the total number of grid cells. This closes the recursion.

\subsection{Relation between PHD/MIB filter and binary Bayes filter}
\label{sec:proof}

The proposed PHD/MIB filter is a generalization of the binary Bayes filter which does not rely on the assumption of a static environment. Consequently, the filter equations should simplify to the well-known equations of the binary Bayes filter for a static process model.

\textit{Proposition:}
Assume a deterministic, static process model in the PHD/MIB filter, so that the predicted intensity distribution for time step $k+1$ is equivalent to the posterior distribution at time $k$, i.e. 
\begin{align}
\Dpred(\xp) = \Dk(\xk) \,\,\, \forall  \,\,\, \xp=\xk \in \stateSpace.
\label{eq:BBF_static_process}
\end{align} 
Further, assume the measurement likelihood $\likecell$ and the clutter density $\pc(\zp)$ are equal uniform distributions in a limited subset $\measSpace_\text{s}$ of the measurement space, i.e.
\begin{align}
\likecell = \pc(\zp) = 
\begin{cases}
\theta  &\text{if } \zp \in \measSpace_\text{s},  \\
0  &\text{otherwise} ,
\end{cases}
\label{eq:constant}
\end{align}
where $\theta>0$ is constant. Then the propagation of the posterior occupancy probability $\exk = \postok^\ci(\occStatek = \occEvent_k)$ at time $k$ to the posterior occupancy probability  $\exkp = \postop^\ci(\occStatep = \occEvent_\kp)$ at time $k+1$ of the PHD/MIB filter reduces to the generalized binary Bayes update \eqref{eq:BBF_gen}
\begin{align}
\postop^\ci(\occEvent_\kp) =
\frac{\alpha^\ci_{\zp}(\occEvent_\kp) \cdot \postok^\ci(\occEvent_k)}
{ \alpha^\ci_{\zp}(\occEvent_\kp) \cdot \postok^\ci(\occEvent_k)
 + \postok^\ci(\freeEvent_\kp)  }, 
\label{eq:BBF_gen_proof}
\end{align}
with
\begin{align}
 \alpha^\ci_{\zp}(\occEvent_\kp) \!=\! 
\begin{cases}
\frac{\cellTPn}{\cellFPn}      	& \!\!\!\! \text{if } \sZp = \emptyset, \\[1.0em]
\frac{\cellTP}{\cellFP}      	&\!\!\!\! \text{if } \sZp = \{ \zp\}.
\end{cases}
\end{align}

Recall that $\occEvent$ and $\freeEvent$ are the two possible cases occupied and free, respectively of the occupancy state $\occState$.

\textit{Proof:}
Due to the static process model \eqref{eq:BBF_static_process}, the Bernoulli distribution for each grid cell does not change during the prediction step:
\begin{align}
\spipred = \spik.
\label{eq:BBF_static_process2}
\end{align} 
In case of no measurement in grid cell $c$, i.e. $\sZp = \emptyset $, the posterior existence probability
of an oject in grid cell $c$ is given by
\begin{align}
\exkp \stackrel{\eqref{eq:occprob_integral}} {=}
&\int\limits_{\xp \in c} \spip  (\{ \xp \}|\emptyset) \,d\xp \nonumber \\
\stackrel{\eqref{eq:filter_update}, \eqref{eq:BBF_static_process2}}{=}  
&\int\limits_{\xp \in c}  \frac{1}{\normempty}
\cdot \cellTPn \cdot 
\spik(\{ \xp \} ) \,d\xp  \nonumber \\
\stackrel{\eqref{eq:normalization_empty}, \eqref{eq:BBF_static_process2}}{=} 
&\frac{ \cellTPn  \cdot  \exk}
{\cellFPn \cdot \exkn + \cellTPn  \cdot  \exk}
\end{align}
and for $\sZp = \{ \zp\} $ it follows
\begin{align}
\exkp \stackrel{\eqref{eq:occprob_integral}}{=} 
&\int\limits_{\xp \in c} \spip  (\{ \xp \}| \{ \zp\}) \,d\xp \nonumber \\
\stackrel{\eqref{eq:filter_update}, \eqref{eq:BBF_static_process2}}{=}  
&\int\limits_{\xp \in c} 
 \frac{\cellTP}{\normz} \cdot  \likecellAs \cdot \spik( \{\xp\} )
 \,d\xp  \nonumber \\
& \!\!\!\!\!\!\!\!\!\!\!\!\!\!\!\!\!\!\!\!\! \stackrel{\eqref{eq:normalization_meas}, \eqref{eq:constant}, \eqref{eq:BBF_static_process2}}{=}  
\frac{ \cellTP \cdot  \likecellAs  \cdot  \exk}
{ \cellFP \cdot \pc \cdot \exkn + \cellTP \cdot \likecellAs \cdot \exk} \nonumber \\
& \!\!\!\!\!\!\!\!\!\!\!\!\! \stackrel{ \eqref{eq:constant}}{=}  
\frac{ \cellTP \cdot  \exk}
{ \cellFP \cdot \exkn + \cellTP \cdot \exk}. \QEDB
\end{align}

\section{Particle Realization of the PHD/MIB Filter}
\label{sec:implementation}

The PHD/MIB filter can be realized in different ways. A main characteristic of the realization is the representation form of the state PHD. This section describes the particle realization of the PHD/MIB filter.

\subsection{Posterior Representation}

Particles are random samples of the posterior PHD at time step $k$. A particle set consists of $\nu$ particles and their weights $\{ \partk, \weightk \}_{i=1}^\nu$. Together they approximate the posterior as
\begin{align}
\Dk(\xk)  \approx \sum_{i=1}^{\nu} \weightk \delta(\xk-\partk).
\end{align}
Here, $\delta$ is the Dirac delta function which satisfies
\begin{align}
\int f(\x) \delta(\x) d\x =  f(0).
\end{align}
\subsection{Prediction of Persistent Objects}
To represent predicted persistent objects, the prediction draws particles by sampling the proposal density $q_\kp$:
\begin{align}
\partperspred \sim q_\kp(\cdot |\partk,\sZp).
\end{align} 
The index $\text{p}$ depicts that these particles represent a persistent point object.
The remaining part of the section assumes it is possible to sample from the transition density $\trans$, which is used as proposal density:
\begin{align}
q_\kp( \x_{\kp} |\partk,\sZp) = \trans( \x_\kp |\partk).
\end{align} 

The sampling provides the predicted particle set 
$\{ \partperspred, 
\weightperspred \}
_{i=1}^{\nu}$
for time step $k+1$, where the particle weights are multiplied with the persistence probability:
\begin{align}
\weightperspred = p_\text{S} \cdot \weightk.
\label{eq:particle_persistence}
\end{align}
The set represents the predicted PHD of persistent point objects:
\begin{align}
\Dpredp(\xp)  \approx \sum_{i=1}^{\nu} \weightperspred \delta(\xp-\partperspred).
\end{align}

\subsection{Transition from a PHD to Multiple Instances of Bernoulli RFSs}

As depicted in Fig. \ref{fig:FilterOverview}, the PHD/MIB filter now transforms the representation form from the joint PHD to individual, independent Bernoulli RFSs for each grid cell. Accordingly, the following steps are carried out for each grid cell $c$ individually. 

Let 
\begin{align}
\{ \partperspredcell, 
\weightperspredcell \}
_{i=1}^{\numParticlesPredCell}
\label{eq:predParticleSet}
\end{align}
be the set of particles predicted into grid cell $c$ at time step $k+1$. The symbol $\numParticlesPredCell$ represents the number of particles predicted into grid cell $c$ at time step $k+1$. 

To keep the notation simple, consider the set \eqref{eq:predParticleSet} as already truncated, i.e., the sum of weights does not exceed 1. If the sum of predicted particle weights in one grid cell exceeds 1, the weights must be normalized to sum up to a number smaller than 1.

According to \eqref{eq:predExistenceProb}, the sum of predicted particle weights in grid cell $c$ then gives the predicted existence probability $\experspred$ of a persistent object in cell $c$:  
\begin{align}
\experspred =  \sum_{i=1}^{\numParticlesPredCell} \weightperspredcell  \,\, \in [0,1] .
\label{eq:predExistenceProbParticles}
\end{align}                                                                                                                                                                                                                                                                                                                                                                                                                                                                                                                                                                                                                                                                                                                                                                                                                                                                                                                                                                                                                                                                                                                                                                                                                                                                                                                                                                                                                                                                                                                                                                                                                                                                                                                                                                                                                                                                                                                                                                                                                                                                                                                                                                                                                                                                                                                                                                                                                                                                                                                                                                                                                                                                                                                                                                                                                                                                                                                                                                                                                                                                                                                                                                                                                                                                                                                                                                                                                                                                                                                                                                                                                                                                                                                                                                                                                                                                                                                                                                                                                                                                                                                                                                                                                                                     %                                                                                                                                                                                                                                                                                                                                                                                                                    
\subsection{Prediction of New-Born Objects}
According to \eqref{eq:predicted_bern}, the predicted existence probability $\exbornpred$ of a new-born object in grid cell $c$ is given by
\begin{align}
\exbornpred = p_\text{B} \cdot \left(1- \experspred \right).
\label{eq:pred_birth_prob}
\end{align} 

Predicted new-born objects in grid cell $c$ are represented by the particle set 
\begin{align}
\{ \partbornpredcell, 
\weightbornpredcell \}
_{i=1}^{\numParticlesBornCell}.
\end{align}

The particles of this set are sampled from the birth distribution:
\begin{align}
\partbornpredcell \sim \pb(\cdot) .
\label{eq:birth_particle_sampling}
\end{align} 
The number of new-born particles $\numParticlesBornCell$ for each grid cell $c$ is a design parameter of the system. It should be chosen individually for each grid cell, depending on the probability of a birth event.

Since the new-born particle weights sum up to the predicted existence probability $\exbornpred$ \eqref{eq:predicted_ex_born} of a new-born object in grid cell $c$, the weight of each new-born particle is given by 
\begin{align}
 \weightbornpredcell  = \frac{\exbornpred}{\numParticlesBornCell} .
\label{eq:predExistenceProbParticlesBorn}
\end{align}                                                                                                                                                                                                                                                                                                                                                                                                                                                                                                                                                                                                                                                                                                                                                                                                                                                                                                                                                                                                                                                                                                                                                                                                                                                                                                                                                                                                                                                                                                                                                                                                                                                                                                                                                                                                                                                                                                                                                                                                                                                                                                                                                                                                                                                                                                                                                                                                                                                                                                                                                                                                                                                                                                                                                                                                                                                                                                                                                                                                                                                                                                                                                                                                                                                                                                                                                                                                                                                                                                                                                                                                                                                                                                                                                                                                                                                                                                                                                                                                                                                                                                                                                                                                                                                    

\subsection{Predicted Bernoulli RFS}

Together, the persistent and the new-born particle sets represent the predicted Bernoulli RFS $\spipred (\sXp)$ in grid cell $c$:
\begin{align}
\spipred (\{ \xp \}) \approx 
   &\sum_{i=1}^{\numParticlesPredCell} \weightperspredcell \delta(\xp - \partperspredcell) \nonumber\\
+  &\sum_{i=1}^{\numParticlesBornCell} \weightbornpredcell \delta(\xp - \partbornpredcell)
\label{eq:predRFSBernoulliParticles}
\end{align}                                                                                                                                                                                                                                                                                                                                                                                                                                                                                                                                                                                                                                                                                                                                                                                                                                                                                                                                                                                                                                                                                                                                                                                                                                                                                                                                                                                                                                                                                                                                                                                                                                                                                                                                                                                                                                                                                                                                                                                                                                                                                                                                                                                                                                                                                                                                                                                                                                                                                                                                                                                                                                                                                                                                                                                                                                                                                                                                                                                                                                                                                                                                                                                                                                                                                                                                                                                                                                                                                                                                                                                                                                                                                                                                                                                                                                                                                                                                                                                                                                                                                                                                                                                                                                                    
and
\begin{align}
\spipred (\emptyset) = 1 - \experspred - \exbornpred.
%\label{eq:predRFSBernoulliEmpty} % Never referenced, so I commented this label which is defined twice.
\end{align}                                                                                                                                                                                                                                                                                                                                                                                                                                                                                                                                                                                                                                                                                                                                                                                                                                                                                                                                                                                                                                                                                                                                                                                                                                                                                                                                                                                                                                                                                                                                                                                                                                                                                                                                                                                                                                                                                                                                                                                                                                                                                                                                                                                                                                                                                                                                                                                                                                                                                                                                                                                                                                                                                                                                                                                                                                                                                                                                                                                                                                                                                                                                                                                                                                                                                                                                                                                                                                                                                                                                                                                                                                                                                                                                                                                                                                                                                                                                                                                                                                                                                                                                                                                                                                                    

\subsection{Particle Update}

Assume a measurement grid map taken at time step $k+1$ provides for each grid cell an observation as stated above in Sect. \ref{sec:phdmib}.
The update step adapts the weights of the particle set \eqref{eq:predRFSBernoulliParticles}. 

%In practical terms, each measurement grid contains the following data: the individual, time-dependent true positive and false positive probabilities $\cellTP$ and $\cellFP$. In case a measurement occurred in the cell it additionally provides the likelihood function $\likecell$ and the corresponding association probability $\pA$.

The update rules for persistent and new-born particles are identical. The notation system uses the weight symbol $w_*$ with $ * \in \{\text{p,b}\}$ in equations that are identical for both persistent particle weights $w_\text{p}$ and new-born particle weights $w_\text{b}$. 

In case a measurement occurred in measurement grid cell $c$,  unnormalized adapted particle weights $\weightunnormcellstar$ are calculated according to \eqref{eq:filter_update}:
\begin{align}
\weightunnormcellstar = \cellTP \cdot \likecellApartstar \cdot \weightpredcellstar.
\label{eq:updateParticleWeightsUnnormalized}
\end{align}                                                                                                                                                                                                                                                                                                                                                                                                                                                                                                                                                                                                                                                                                                                                                                                                                                                                                                                                                                                                                                                                                                                                                                                                                                                                                                                                                                                                                                                                                                                                                                                                                                                                                                                                                                                                                                                                                                                                                                                                                                                                                                                                                                                                                                                                                                                                                                                                                                                                                                                                                                                                                                                                                                                                                                                                                                                                                                                                                                                                                                                                                                                                                                                                                                                                                                                                                                                                                                                                                                                                                                                                                                                                                                                                                                                                                                                                                                                                                                                                                                                                                                                                                                                                                                                    
The normalized weights are given by 
\begin{align}
\weightpcellstar = \frac{\weightunnormcellstar}{\normz}
\label{eq:applyNormalization}
\end{align}                                                                                                                                                                                                                                                                                                                                                                                                                                                                                                                                                                                                                                                                                                                                                                                                                                                                                                                                                                                                                                                                                                                                                                                                                                                                                                                                                                                                                                                                                                                                                                                                                                                                                                                                                                                                                                                                                                                                                                                                                                                                                                                                                                                                                                                                                                                                                                                                                                                                                                                                                                                                                                                                                                                                                                                                                                                                                                                                                                                                                                                                                                                                                                                                                                                                                                                                                                                                                                                                                                                                                                                                                                                                                                                                                                                                                                                                                                                                                                                                                                                                                                                                                                                                                                                    
with \eqref{eq:normalization_meas}
\begin{align}
\normz & \approx 
\cellFP \cdot \pc(\zp) \cdot \expredn \nonumber \\
&  + \sum_{i=1}^{\numParticlesPredCell} \weightunnormperscell 
+ \sum_{i=1}^{\numParticlesBornCell} \weightunnormborncell .
\label{eq:normalizationCalculation}
\end{align}                                                                                                                                                                                                                                                                                                                                                                                                                                                                                                                                                                                                                                                                                                                                                                                                                                                                                                                                                                                                                                                                                                                                                                                                                                                                                                                                                                                                                                                                                                                                                                                                                                                                                                                                                                                                                                                                                                                                                                                                                                                                                                                                                                                                                                                                                                                                                                                                                                                                                                                                                                                                                                                                                                                                                                                                                                                                                                                                                                                                                                                                                                                                                                                                                                                                                                                                                                                                                                                                                                                                                                                                                                                                                                                                                                                                                                                                                                                                                                                                                                                                                                                                                                                                                                                    
In case no measurement occurred in measurement grid cell $c$, the update rule for both the persistent and new-born particles to calculate adapted weights $\weightpcellstar$ is according to \eqref{eq:filter_update}:

\begin{align}
\weightpcellstar = 
\frac{\cellTPn}
{\cellTPn  \cdot \expred + \cellFPn \cdot \expredn} 
\weightpredcellstar.
\label{eq:updateParticleWeightsEmpty}
\end{align}                                                                                                                                                                                                                                                                                                                                                                                                                                                                                                                                                                                                                                                                                                                                                                                                                                                                                                                                                                                                                                                                                                                                                                                                                                                                                                                                                                                                                                                                                                                                                                                                                                                                                                                                                                                                                                                                                                                                                                                                                                                                                                                                                                                                                                                                                                                                                                                                                                                                                                                                                                                                                                                                                                                                                                                                                                                                                                                                                                                                                                                                                                                                                                                                                                                                                                                                                                                                                                                                                                                                                                                                                                                                                                                                                                                                                                                                                                                                                                                                                                                                                                                                                                                                                                                    

Notice that for  multi-object distributions, normalization does not mean all particle weights sum up to 1. Instead, the sum of updated particle weights equals the posterior existence probability $\exkp$ of a point object in grid cell $c$ at time step $k+1$, which is also the posterior occupancy probability $\postop^\ci(\occEventp)$ of the grid cell: 
\begin{align}
\exkp = \postop^\ci(\occEventp) =
\sum_{i=1}^{\numParticlesPredCellp} \weightperspcell 
+ \sum_{i=1}^{\numParticlesBornCellp} \weightbornpcell.
\label{eq:occupancyProbParticles}
\end{align}                                                                                                                                                                                                                                                                                                                                                                                                                                                                                                                                                                                                                                                                                                                                                                                                                                                                                                                                                                                                                                                                                                                                                                                                                                                                                                                                                                                                                                                                                                                                                                                                                                                                                                                                                                                                                                                                                                                                                                                                                                                                                                                                                                                                                                                                                                                                                                                                                                                                                                                                                                                                                                                                                                                                                                                                                                                                                                                                                                                                                                                                                                                                                                                                                                                                                                                                                                                                                                                                                                                                                                                                                                                                                                                                                                                                                                                                                                                                                                                                                                                                                                                                                                                                                                                    

The posterior Bernoulli RFS $\spip (\sXp)$ of grid cell $c$ is given by:
\begin{align}
\spip (\{ \xp \}) \approx 
   &\sum_{i=1}^{\numParticlesPredCellp} \weightperspcell \delta(\xp - \partperspcell) \nonumber\\
+  &\sum_{i=1}^{\numParticlesBornCellp} \weightbornpcell \delta(\xp - \partbornpcell)
\label{eq:postRFSBernoulliParticles}
\end{align}                                                                                                                                                                                                                                                                                                                                                                                                                                                                                                                                                                                                                                                                                                                                                                                                                                                                                                                                                                                                                                                                                                                                                                                                                                                                                                                                                                                                                                                                                                                                                                                                                                                                                                                                                                                                                                                                                                                                                                                                                                                                                                                                                                                                                                                                                                                                                                                                                                                                                                                                                                                                                                                                                                                                                                                                                                                                                                                                                                                                                                                                                                                                                                                                                                                                                                                                                                                                                                                                                                                                                                                                                                                                                                                                                                                                                                                                                                                                                                                                                                                                                                                                                                                                                                                    
and
\begin{align}
\spip (\emptyset) = \exkpn = 1 - \exkp.
%\label{eq:predRFSBernoulliEmpty} % Never referenced, so I commented this label which is defined twice.
\end{align}

\subsection{Joint PHD Representation}
The PHD/MIB represents the posterior multi-object state of all point objects in the environment by its PHD. 
The transformation from multiple instances of Bernoulli RFSs to a joint PHD is given by \eqref{eq:jointPDH}
\begin{align}
\Dp(\xp)  \approx 
\!\!\!\!\!\!\!\!   &\sum_{i \in [1,\numParticlesPredCellp], c \in [1,C] } \!\!\!\!\!\!\!\! \weightperspcell \delta(\xp - \partperspcell) \nonumber\\
+\!\!\!\!\!\!\!\!   &\sum_{i \in [1,\numParticlesBornCellp], c \in [1,C] } \!\!\!\!\!\!\!\! \weightbornpcell \delta(\xp - \partbornpcell).
\label{eq:postPHDParticles}
\end{align}                                                                                                                                                                                                                                                                                                                                                                                                                                                                                                                                                                                                                                                                                                                                                                                                                                                                                                                                                                                                                                                                                                                                                                                                                                                                                                                                                                                                                                                                                                                                                                                                                                                                                                                                                                                                                                                                                                                                                                                                                                                                                                                                                                                                                                                                                                                                                                                                                                                                                                                                                                                                                                                                                                                                                                                                                                                                                                                                                                                                                                                                                                                                                                                                                                                                                                                                                                                                                                                                                                                                                                                                                                                                                                                                                                                                                                                                                                                                                                                                                                                                                                                                                                                                                                                    

Usually, particle filter realizations of a PHD filter provide only the persistent part of the posterior PHD as output \cite{Ristic2013c}. Depending on the application, new-born particles considerably increase the uncertainty of the estimated state of objects. So it is often beneficial to consider their influence on the state estimation only after another recursion.

\subsection{Resampling}
For many applications it is important to keep the overall number of used particles constant. Therefore, the PHD/MIB filter resamples the constant number of $\nu$ particles from the joint posterior particle set. For each particle, the probability to be drawn is proportional to its weight. Let $\{ \x_\kp^\ind, w_\kp^\ind \}_{i=1}^\nu$ be the set of resampled particles and their weights. 
The new weights of the particles are all equal and normalized to sum up to the same value as the posterior weights of the persistent and the new-born particles together: 
% \textbf{do we really need this equation?}
\begin{align}
\sum_{i \in [1,\nu] } \!\!\!\! w_\kp^{\ind} \,\,\,\,\,
= \!\!\!\!  &\sum_{i \in [1,\numParticlesPredCellp], c \in [1,C] } \!\!\!\!\!\!\!\! \weightperspcell  \nonumber\\
+ \!\!\!\!  &\sum_{i \in [1,\numParticlesBornCellp], c \in [1,C] } \!\!\!\!\!\!\!\! \weightbornpcell. 
\label{eq:resampledParticleWeights}
\end{align}                                                                                                                                                                                                                                                                                                                                                                                                                                                                                                                                                                                                                                                                                                                                                                                                                                                                                                                                                                                                                                                                                                                                                                                                                                                                                                                                                                                                                                                                                                                                                                                                                                                                                                                                                                                                                                                                                                                                                                                                                                                                                                                                                                                                                                                                                                                                                                                                                                                                                                                                                                                                                                                                                                                                                                                                                                                                                                                                                                                                                                                                                                                                                                                                                                                                                                                                                                                                                                                                                                                                                                                                                                                                                                                                                                                                                                                                                                                                                                                                                                                                                                                                                                                                                                                    

\section{Real-time Approximation with Dempster-Shafer Theory of Evidence}
\label{sec:ds_appr}

For some application scenarios, the presented particle realization of the PHD/MIB filter might not be real-time capable. A possible reason are huge unobserved areas in grid maps. Since the presented particle realization of the PHD/MIB filter represents potential point objects in unobserved areas with particles, it requires a large number of them. 
All mentioned publications of particle-based dynamic grid maps \cite{Danescu2011, Tanzmeister2014, Negre2014, Nuss2015} use particles only for occupied grid cells, not for unobserved grid cells. 
One possibility to distinguish between unobserved and occupied cells is to use Dempster-Shafer masses of evidence \cite{Dempster1968}, \cite{Shafer1976}, \cite{Smets1990} instead of occupancy probabilities. Both \cite{Tanzmeister2014} and \cite{Nuss2015} create particles only in areas with evidence for occupancy.

This section presents a coarse approximation of the particle-based PHD/MIB filter, applying the Dempster-Shafer theory of evidence. The resulting approximation will be referred to as DS-PHD/MIB filter. The DS-PHD/MIB filter is able to run with a substantially reduced number of particles compared to the original PHD/MIB filter and is also easier to implement. An efficient, massively parallelized implementation of the DS-PHD/MIB filter will be presented in Sect. \ref{sec:parallel_implementation}.

\subsection{State Representation}
An introduction to the Dempster-Shafer theory of evidence and grid maps can be found in \cite{Nuss2013, Konrad2011, Moras2011}.
The DS-PHD/MIB filter represents the occupancy state of a grid cell with a basic belief assignment (BBA) $m: 2^\Omega \rightarrow [0,1]$. The frame of discernment 
$\Omega$ contains the events occupied and free: $\Omega = \{\occEvent,\freeEvent\}$. So each grid cell stores a mass for occupied $m(\occEvent)$ and a mass for free $m(\freeEvent)$. The propagation of these masses over time are carried out separately. The propagation of the mass for free is estimated as in a static grid map. The propagation of the mass for occupied is motivated by the PHD/MIB filter. 

The DS-PHD/MIB filter represents the posterior state of an individual grid cell $c$ at time $k$ with the particle set
$\{ \partkcell, \weightkcell \}_{i=1}^{\numParticlesKCell}$
and the mass for free $\bbakcell(\freeEvent_k)$. Here, the sum of particle weights represents the mass for occupied: 
\begin{align}
\bbakcell(\occEvent_k) = \sum_{i=1}^{\numParticlesKCell} \weightkcell. 
\label{eq:DS_particle_sum}
\end{align}
The occupancy probability $\postok^\ci(\occEvent)$ in a grid cell is given by the pignistic transformation
\begin{align}
\postok^\ci(\occEvent_k) = \bbakcell(\occEvent_k) + 0.5 \cdot (1 - \bbakcell(\occEvent_k) - \bbakcell(\freeEvent_k)). 
\end{align}
The distribution of the particles approximates the spatial PDF $p_k^\ci(x_k)$ of a point object in grid cell $c$:
\begin{align}
p_k^\ci(x_k)  \approx \frac{1}{\bbakcell(\occEvent_k)} \sum_{i=1}^{\numParticlesKCell} \weightkcell \delta(x_k-\partkcell). 
\label{eq:DS_pdf}
\end{align}

\begin{figure*}[t]
  \centering
  \includegraphics[width=0.7\textwidth]{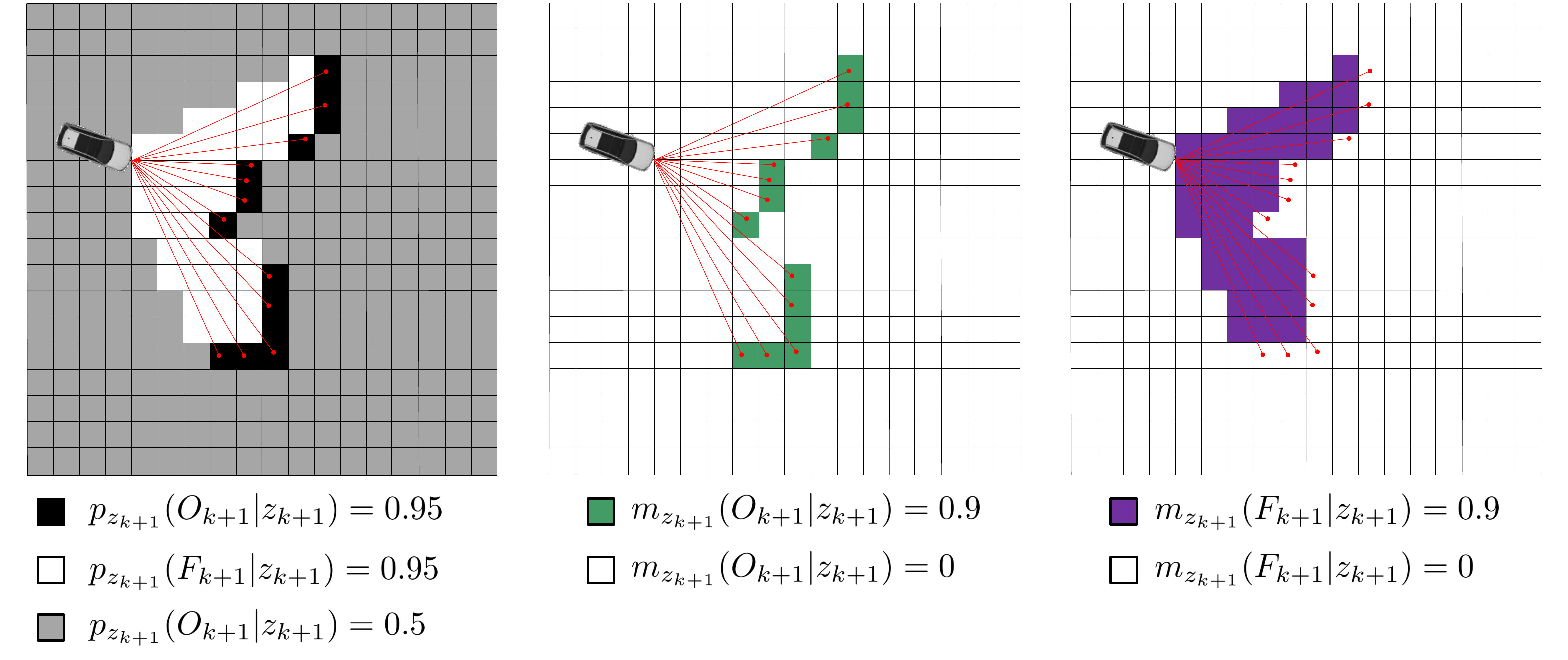}
	\caption{A simple occupancy measurement grid originated by a laser measurement with multiple beams. The red lines represent laser beams that hit obstacles at their ends. The grid on the left-hand side provides an occupancy probability $p_{\zp}(\occStatep|\zp)$ for each grid cell. The middle figure and the figure on the right-hand side show the same occupancy measurement grid represented by evidences for occupied $\bbazp(\occEventp|\zp)$ (middle) and free $\bbazp(\freeEventp|\zp)$ (right), respectively. The pignistic transformation of the Dempster-Shafer grid results in the classical grid on the left-hand side \cite{Nuss2015}. The DS-PHD/MIB filter initializes new particles only in green grid cells.}
	\label{fig:DSMeasGrid}
\end{figure*}

\subsection{Prediction}
The DS-PHD/MIB filter applies \eqref{eq:process} and \eqref{eq:particle_persistence} to predict particles to the next time step. In analogy to the PHD/MIB filter \eqref{eq:predParticleSet}, let 
$\{ \partperspredcell, 
\weightperspredcell \}
_{i=1}^{\numParticlesPredCell}
$
be the set of particles predicted into cell $c$ at time step $k+1$. Again, the predicted weights are truncated, so the sum of predicted weights in one grid cell is limited to 1. 

The DS-PHD/MIB filter estimates the predicted occupancy mass of grid cell $c$ by
\begin{align}
\bbapredcell(\occEvent_\kp) = \sum_{i=1}^{\numParticlesPredCell}  \weightperspredcell.  
\label{eq:DS_particle_sum_pred}
\end{align}

The predicted mass for free is modeled as in a static grid map and given by 
\begin{align}
\label{eq:DS_freespace_pred}
\bbapredcell(\freeEvent_\kp) = \, &\text{min}\left[ \alpha(T) \bbakcell(\freeEvent_k) \, , \,\, 1-\bbapredcell(\occEvent_\kp)    \right] ,
\end{align}
where the discount factor $\alpha(T) \in [0,1]$ models the decreasing prediction reliability, depending on the time interval $T$ between two update steps. Since the sum of masses cannot exceed $1$, the predicted free space evidence is limited accordingly.  

\subsection{Update}
The PHD/MIB filter considers both the existence probability and the spatial distribution of point objects in a joint Bayesian innovation step, formally derived as a Bernoulli filter. The DS-PHD/MIB filter does not formally derive the update step, but uses heuristically designed, simplified update equations with the goal of modeling the probabilistic update equations of the PHD/MIB filter as close as possible in the Dempster-Shafer domain.

The DS-PHD/MIB approximation updates the existence probability of a point object in grid cell $c$ independently of its spatial distribution. 
Accordingly, the DS-PHD/MIB filter expects the following information in each measurement grid cell:
\begin{itemize}
\item The observed occupancy BBA \mbox{$\bbazpcell: 2^{\{\occEvent,\freeEvent\}} \rightarrow [0,1]$},
\item the spatial likelihood function $\likecell$, and
\item the association probability $\pA$ between the likelihood function $\likecell$ and the point object.
\end{itemize} 
Figure \ref{fig:DSMeasGrid} shows an example for a measurement grid with occupancy BBAs. An example for a likelihood function $\likecell$ and the association probability $\pA$ in measurement grid cells can be found in \cite{Nuss2015}, where it results from radar doppler measurements.

\subsubsection{Existence Update}
The DS-PHD/MIB filter approximates the existence update by simply combining the predicted BBF $\bbapredcell$ and the observed BBA $\bbazpcell$ of the corresponding measurement grid cell with the Dempster-Shafer rule of combination (see \cite{Nuss2013}):
\begin{align}
\bbakpcell = \bbapredcell \oplus \bbazpcell.
\label{eq:DS_comb}
\end{align}
\subsubsection{Birth Model}
The DS-PHD/MIB filter splits the mass for occupied into two parts: occupied by a persistent object and occupied by a new-born object, denoted as:
\begin{align} 
\bbakpcell(\occEvent_\kp) = \massperskpcell + \massbornkpcell.
\label{eq:DS_sum_of_mass}
\end{align}

Assume the PHD/MIB filter updates the state of a point object with a uniformly distributed likelihood $\likecell$. Then the relation between the updated existence probability $\exbornkp$ of a new-born object and the updated existence probability $\experskp$ of a persistent object results in 
%todo \textbf{Todo: Derivation in Appendix}
\begin{align} 
\frac{\exbornkp}{\experskp} =
\frac{\exbornpred}{\experspred}   = 
\frac{\pB \left[1- \experspred\right]}{\experspred}.
\label{eq:DS_relation_probabilities}
\end{align}
Analogously, the DS-PHD/MIB models the relation between masses for a new-born and a persistent object as
\begin{align} 
\frac{\massbornkpcell}{\massperskpcell} = 
\frac{\pB \left[1 - \massperskpcell  \right]}{\massperskpcell}.
\label{eq:DS_relation_masses}
\end{align}
Combining \eqref{eq:DS_sum_of_mass} and \eqref{eq:DS_relation_masses} delivers the resulting masses for a new-born and a persistent object:
\begin{align} 
&\massbornkpcell =  
\frac{ \bbakpcell(\occEvent_\kp) \cdot \pB \left[1-  \bbapredcell(\occEvent_\kp)\right] }
{\bbapredcell(\occEvent_\kp)  + \pB \left[1-  \bbapredcell(\occEvent_\kp)\right]},  \label{eq:DS_resulting_masses1}\\
&\massperskpcell = \bbakpcell(\occEvent_\kp) -\massbornkpcell.
\label{eq:DS_resulting_masses2}
\end{align}

\subsubsection{Spatial Update}
The DS-PHD/MIB filter provides three particle sets to approximate the posterior spatial distribution $p^\ci_\kp(\xp)$. 
The first particle set represents a persistent object and results from the set predicted into grid cell $c$, denoted as
$\{ \partperspredcell, 
\weightperspredcell \}
_{i=1}^{\numParticlesPredCell}
$. It is updated by multiplying the weights with the spatial measurement likelihood $\likecell$. This leads to the unnormalized updated weights 
\begin{align}
\weightunnormperscell =  \likecellpart \cdot \weightperspredcell.
\label{eq:DSupdateParticleWeightsUnnormalized}
\end{align} 
The particle states remain unchanged: 
\begin{align}
\partperspcell = \partperspredcell.
\label{eq:DSparticle_states}
\end{align}
The normalized particle weights are given by 
\begin{align}
\weightperspcell = \pA \!\cdot\! \mu^\ci_\text{A}  \!\cdot\! \weightunnormperscell + 
\left( 1- \pA \right) \!\cdot\! \mu^\ci_{\overline{\text{A}}}  \!\cdot\! \weightperspredcell ,
\label{eq:DSnormalization}
\end{align}
with
\begin{align}
\mu^\ci_\text{A} = \left[ \sum_{i=1}^{\numParticlesPredCell} \weightunnormperscell\right]^{-1} 
\!\!\ \cdot \,
 \, &\massperskpcell  
\label{eq:DSnormalizationfactor}
\end{align}
and
\begin{align}
\mu^\ci_{\overline{\text{A}}} = \left[ \sum_{i=1}^{\numParticlesPredCell} \weightperspredcell
\right]^{-1} \!\!\ \cdot \,
 \, &\massperskpcell
  = \frac{\massperskpcell}{\bbapredcell(\occEvent_\kp)}.
\label{eq:DSnormalizationfactor2}
\end{align}
Equation \eqref{eq:DSnormalization} considers that with a probability of $(1-\pA)$, the likelihood function $\likecellpart$ is not associated with the point object in the grid cell. 
In this case, the weight update and normalization step serves solely to normalize the predicted particle weights in such a way that they sum up to the posterior persistent occupancy pass $\massperskpcell$.

The second and third particle sets represent new-born objects. For computational efficiency reasons, they are only created in grid cells where the corresponding measurement grid cell reports a mass for occupied: $\bbazpcell(\occEvent_\kp)>0$.
The second particle set 
$\{ \partasspcell, 
\weightasspcell \}
_{i=1}^{\numParticlesAssCellp}
$
represents a new-born object under the assumption that the spatial measurement $z^\ci_\kp$ in grid cell $c$ is associated to the point object in grid cell $c$. The particles are sampled from the probability density function $p^\ci_{x_\kp}(x_\kp|z^\ci_\kp)$ of the state $x_\kp$ given the measurement $z_\kp^\ci$ in grid cell $c$:
\begin{align}
\partasspcell \sim p^\ci_{x_\kp}(\cdot|z^\ci_\kp).
\label{eq:DSbirth_particle_samplingA}
\end{align} 
The weight of each particle in the second set can directly be calculated to 
\begin{align}
\weightasspcell = \frac{\pA \cdot \massbornkpcell}{\numParticlesAssCellp}.
\label{eq:weights_associated} 
\end{align}

Details about creating a probability density function $p^\ci_{x_\kp}(x_\kp|z^\ci_\kp)$ of the state $x_\kp$ given the measurement $z^\ci_\kp$ can be found in \cite{Ristic2013c}, p.~38.

The third particle set 
$\{ \partassnpcell, 
\weightassnpcell \}
_{i=1}^{\numParticlesAssNCellp}
$
represents a new-born object under the assumption that the spatial measurement $z^\ci_\kp$ in grid cell $c$ is not associated to the point object in grid cell $c$. The particles of this set are sampled from the birth distribution $\birthp(\xp)$:
\begin{align}
\partassnpcell \sim \birthp(\cdot) 
\label{eq:DSbirth_particle_sampling}
\end{align} 
The weight of each particle in the third set can directly be calculated to 
\begin{align}
\weightassnpcell = \frac{\left(1- \pA\right) \cdot \massbornkpcell}{\numParticlesAssNCellp}.
\label{eq:weights_unassociated} 
\end{align} 

When creating the second and the third particle set, the individual particle numbers $\numParticlesAssCellp$ and $\numParticlesAssNCellp$ of each grid cell should relate to their corresponding occupancy masses.

Finally, the posterior spatial state distribution of the point object in grid cell $c$ at time $k+1$ is given by 
\begin{align}
p_\kp^\ci(x_\kp)  \approx \frac{1}{\bbakpcell(\occEvent_\kp)} \sum_{i=1}^{\numParticlesPCell} \weightpcell \delta(x_\kp-\partpcell),
\label{eq:DS_pdf_post}
\end{align}
where the set 
$\{ \partpcell, 
\weightpcell \}
_{i=1}^{\numParticlesPCell}
$
is the union of all three particle sets created in the spatial update step. The grid cell additionally stores the posterior mass for free $\bbakpcell(\freeEvent_\kp)$ as calculated in \eqref{eq:DS_comb}, which completes the posterior state together with the particle set. 

\subsection{Resampling}
The resampling step is identical to the resampling step of the original PHD/MIB filter.

\section{Parallel Implementation}
\label{sec:parallel_implementation}

This section describes an implementation of the particle-based DS-PHD/MIB filter with a focus on massively parallel processing systems such as graphics processing units. 

\subsection{Parallelization Challenges}
Particles can naturally be processed in parallel, but here a challenge is to assign each particle to its corresponding grid cell in an efficient way. The assignment is necessary to predict the grid cell occupancy mass \eqref{eq:DS_particle_sum_pred} and to calculate the normalization factor during the update step \eqref{eq:DSnormalization}. Another challenge is to calculate statistical moments of grid cells as mean and variance of the velocity in a balanced way: the calculation time should be independent of the number of particles assigned to a grid cell.

The proposed solution sorts the particles after the prediction step according to the grid cell index they have been predicted into.
Sorting particles has a time complexity quasilinear in the number of particles.
Although the parallelization potential of sorting is somewhat limited, there are sophisticated sorting algorithms capable of achieving a high throughput especially on massively parallel architectures \cite{Satish2009}.
The availability of a sorted particle array has several advantages:
First, the assignment of sorted particles to grid cells is straightforward as will be detailed below.
Furthermore, particle state values can be efficiently accumulated in separate arrays, similar to an integral image data structure \cite{Viola2001}.
This facilitates calculation of a grid cell's statistical moments with a computational complexity independent of the number of particles assigned to the cell.

Another advantage of sorting particles is a simple overall implementation since all remaining advanced problems can then be solved with standard routines.
Highly efficient parallel implementations of these routines are available for graphics processing units, where for the parallel implementation of the particle-based DS-PHD/MIB filter sampling of random numbers \cite{Nvidia2016}, sorting \cite{Hoberock2016} and accumulation have been employed.

\subsection{Implementation Details}
\begin{figure}[t]
  \centering
  \includegraphics[width=0.67\columnwidth]{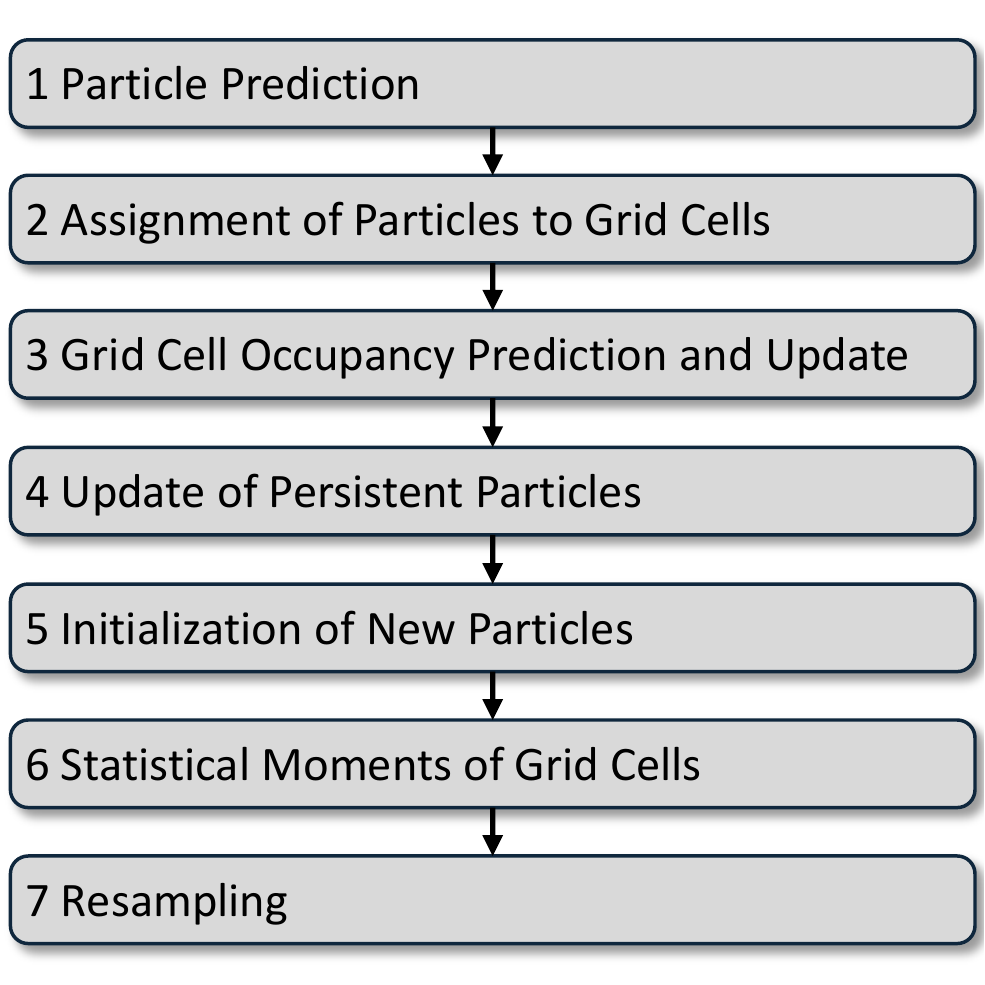}
	\caption{ Overview of the PHD/MIB implementation, broken down into seven steps.}
	\label{fig:AlgorithmOverview}
\end{figure}

In the following, implementation details of the proposed parallel algorithm are described.
The auxiliary data structures rendering the algorithm particularly efficient are given as follows.
All particles and grid cells are stored in the $\textit{particle\_array}$ and $\textit{grid\_cell\_array}$ arrays, respectively.
Assume a measurement grid map with the same dimensions as the grid map is already available and measurement grid cells are stored in the array $\textit{meas\_cell\_array}$.
Details about efficient measurement grid calculation can be found in \cite{Homm2010}.
For modeling noise processes, a sufficient amount of random numbers is sampled beforehand during idle times and stored in extra arrays.
The parallel DS-PHD/MIB recursion is summarized in Fig. \ref{fig:AlgorithmOverview} and outlined in the following sections. 

\subsubsection{Particle Prediction}
Algorithm \ref{al:particle_prediction} provides pseudo code for the particle prediction. The algorithm predicts all particles in parallel applying equations \eqref{eq:process} and \eqref{eq:particle_persistence}. This includes calculating the new grid cell index of each particle after prediction. The appropriate amount of random numbers should be sampled in a separate step in advance, so the prediction step just needs to look them up.
\begin{algorithm*}[p]
\caption{\small Particle Prediction}\label{al:particle_prediction}
\begin{algorithmic}[1]\footnotesize
\State $\textit{particle\_array}$  \Comment{This array stores particles including weight and corresponding grid cell index (constant size $\nu$)}
\State $\textit{grid\_cell\_array}$  \Comment This array stores the grid cells (constant size $C$)
\State $\textit{p\_S}$	\Comment{Persistence probability of point objects, is a design parameter of the 
process model}
\For {$i \in \{0,\dotsc,\text{length}(\textit{particle\_array})-1 \}$} \Comment{Parallel for loop over all particles}
	\State $\text{predict}(\textit{particle\_array},i,p\_S)$ 
	\Comment{Applies \eqref{eq:process} and \eqref{eq:particle_persistence}, calculates new grid cell index and stores it inside the particle}
\EndFor
\end{algorithmic}
\end{algorithm*}

\subsubsection{Assignment of Particles to Grid Cells}
Pseudo code for the particle assignment is given in Algorithm \ref{al:assign_particles}. First, the algorithm sorts all particles according to the grid cell index they have been predicted into. 
Each grid cell can store two particle indices. They represent the first and last index of the particle group that has been predicted into the grid cell. For the assignment, each particle checks if it is the first or the last particle of a group with the same index. If so, it writes its index into the according grid cell. Since there can only be at most one first or last particle per grid cell, the assignment can run in parallel without any writing conflicts. 
\begin{algorithm*}[p]
\caption{\small Assignment of Particles to Grid Cells}\label{al:assign_particles}
\begin{algorithmic}[1]\footnotesize
\State $\textit{weight\_array}$  \Comment{This array will be an additional storage for the particle weights (constant size $\nu$)}
\State $\text{sort}(\textit{particle\_array})$  \Comment{Sorts by the grid cell index the particle has been predicted into}
\For {$i \in \{0,\dotsc,\text{length}(\textit{particle\_array})-1 \}$} \Comment{Parallel for loop over all particles}
	\State $j \gets \text{get\_grid\_cell\_idx}(\textit{particle\_array}[i])$ \Comment{Reads the grid cell index of the predicted particle}
	\If {$\text{is\_first\_particle}(\textit{particle\_array},i)$} \Comment{Checks if $i$ is the first particle of a group with same grid cell index}
		\State $\text{set\_particle\_start\_idx}(\textit{grid\_cell\_array},j,i)$ \Comment{Sets $i$ as particle start index of grid cell $j$ in \textit{grid\_cell\_array}}
	\EndIf
	\If {$\text{is\_last\_particle}(\textit{particle\_array},i)$} \Comment{Checks if $i$ is the last particle of a group with same grid cell index}
		\State $\text{set\_particle\_end\_index}(\textit{grid\_cell\_array},j,i)$ \Comment{Sets $i$ as particle end index of grid cell $j$ in \textit{grid\_cell\_array}}
	\EndIf
	\State $\textit{weight\_array}[i] \gets \text{get\_particle\_weight}(\textit{particle\_array}[i])$ \Comment{Copies weight of particle $i$ to \textit{weight\_array} at index $i$}
\EndFor
\end{algorithmic}
\end{algorithm*}

\subsubsection{Grid Cell Occupancy Prediction and Update}
Algorithm \ref{al:existence_prediction} details the occupancy update. The goal of this step is to calculate for each grid cell the predicted and updated occupancy BBA.
First, the algorithm accumulates in parallel all particle weights and stores the result in the array $\textit{weight\_array\_accum}$. The remaining part of the algorithm is carried out in parallel for all grid cells. Each cell reads two values from $\textit{weight\_array\_accum}$. The first value is the accumulated particle weight of all preceding grid cells excluding its own weight, the second value is the accumulated particle weight of all preceding grid cells including its own weight. The cell simply subtracts the first value from the second value to calculate its predicted occupancy mass according to \eqref{eq:DS_particle_sum_pred} with constant time complexity. The predicted free mass is calculated according to \eqref{eq:DS_freespace_pred}. 

Each cell reads the observed occupancy BBA from the corresponding measurement grid cell and combines it with its predicted occupancy BBA according to \eqref{eq:DS_comb} to calculate its updated occupancy BBA. This enables the grid cell to separate the posterior occupancy mass $\bbakpcell (\occEvent_\kp)$ into the part $\massperskpcell$ for a persistent object and the part $\massbornkpcell$ for a new-born object. Each cell stores the part $\massbornkpcell$ for a new-born object in a separate array, which will be used later to calculate the number of newly drawn particles for this cell.
\begin{algorithm*}[p]
\caption{\small Grid Cell Occupancy Prediction and Update}\label{al:existence_prediction}
\begin{algorithmic}[1]\footnotesize
\State $\textit{meas\_cell\_array}$  \Comment This array stores the measurement grid cells (constant size $C$)
\State $\textit{born\_masses\_array}$  \Comment{This array will store the mass for a new-born object per grid cell (constant size $C$)}
\State $\textit{p\_B}$ \Comment{Birth probability of point objects, is a design parameter of the process model}
\State $\textit{weight\_array\_accum} \gets \text{accumulate}(\textit{weight\_array})$   \Comment{Inclusively accumulates all particle weights to \textit{weight\_array\_accum}}
\For {$j \in \{0,\dotsc,\text{length}(\textit{grid\_cell\_array})-1 \}$} \Comment{Parallel for loop over all grid cells}
	\State $\textit{start\_idx} \gets \text{get\_particle\_start\_idx}(\textit{grid\_cell\_array}[j])$  \Comment{Gets start index in \textit{particle\_array} of cell $j$ }
	\State $\textit{end\_idx} \gets \text{get\_particle\_end\_idx}(\textit{grid\_cell\_array}[j])$ \Comment{Gets end index in \textit{particle\_array} of cell $j$}
	\State $\textit{m\_occ\_pred} \gets \text{subtract}(\textit{weight\_array\_accum}, \textit{start\_idx}, \textit{end\_idx})$ \Comment{Calculates predicted occupancy mass of  cell $j$ \eqref{eq:DS_particle_sum_pred}}
	\State $\textit{m\_free\_pred} \gets \text{predict\_free\_mass}( \textit{grid\_cell\_array}[j])$ \Comment{Predicts free mass of cell $j$ \eqref{eq:DS_freespace_pred}}
	\State $\textit{m\_occ\_up} \gets \text{update\_o}(\textit{m\_occ\_pred}, \textit{m\_free\_pred}, \textit{meas\_cell\_array}[j])$ \Comment{Combination to posterior occ. mass \eqref{eq:DS_comb}}
	\State $\textit{m\_free\_up} \gets \text{update\_f}(\textit{m\_occ\_pred}, \textit{m\_free\_pred}, \textit{meas\_cell\_array}[j])$ \Comment{Combination to posterior free mass \eqref{eq:DS_comb}}
	\State $\textit{rho\_b} \gets \text{separate\_newborn\_part}(\textit{m\_occ\_pred}, \textit{m\_occ\_up}, p\_B)$ \Comment{Calculate new-born part of posterior occupancy mass \eqref{eq:DS_resulting_masses1}}
	\State $\textit{rho\_p} \gets \textit{m\_occ\_up} - \textit{rho\_b}$ \Comment{Calculates remaining persistent part of posterior occupancy mass \eqref{eq:DS_resulting_masses2}}
	\State $ \textit{born\_masses\_array}[j] \gets \textit{rho\_b}$ \Comment{Stores new-born part of posterior occupancy mass of cell $j$ in \textit{born\_masses\_array}}
	\State $\text{store\_values}(\textit{rho\_b}, \textit{rho\_p}, \textit{m\_free\_up}, \textit{grid\_cell\_array}, j)$ \Comment{Stores updated BBA in grid cell $j$}

\EndFor
\end{algorithmic}
\end{algorithm*}

\subsubsection{Update of Persistent Particles}
The update of persistent particles is described in Algorithm \ref{al:update}. Each particle has stored its corresponding grid cell index during the prediction step, which is assumed the same as the corresponding measurement grid cell index. All persistent particles calculate in parallel their unnormalized updated weight according to \eqref{eq:DSupdateParticleWeightsUnnormalized}. These weights are then accumulated in the array $\textit{weight\_array\_accum}$. Recall that each grid cell has already stored the index range of its corresponding particles in Algorithm \ref{al:assign_particles}. Consequently, in a parallel \textbf{\footnotesize for} loop, each grid cell can look up the accumulated weight of its updated, unnormalized particles in $\textit{weight\_array\_accum}$ analogously to Algorithm \ref{al:existence_prediction}. Each grid cell uses the result to calculate its normalization component $\mu^\ci_\text{A}$ \eqref{eq:DSnormalizationfactor} and stores it. The other normalization component $\mu^\ci_{\overline{\text{A}}}$ as given by \eqref{eq:DSnormalizationfactor2} can directly be calculated and stored in the grid cell. In the next parallel \textbf{\footnotesize for} loop over all particles, each particle uses the grid map as a lookup table for its normalization components $\mu^\ci_\text{A}$ and $\mu^\ci_{\overline{\text{A}}}$ and normalizes itself \eqref{eq:DSnormalization}.
\begin{algorithm*}[p]
\caption{\small Update of Persistent Particles}\label{al:update}
\begin{algorithmic}[1]\footnotesize
\For {$i \in \{0,\dotsc,\text{length}(\textit{particle\_array})-1 \}$} \Comment{Parallel for loop over all persistent particles}
	\State $ \textit{weight\_array}[i] \gets  \text{update\_unnorm}(\textit{particle\_array},i, \textit{meas\_cell\_array})$ 
	\Comment{Calculates unnormalized weight update  \eqref{eq:DSupdateParticleWeightsUnnormalized}}
\EndFor
\State $\textit{weight\_array\_accum} \gets \text{accumulate}(\textit{weight\_array})$   \Comment{Accumulates unnormalized weights of persistent particles}
\For {$j \in \{0,\dotsc,\text{length}(\textit{grid\_cell\_array})-1 \}$ } \Comment{Parallel for loop over all grid cells}
	\State $\textit{start\_idx} \gets \text{get\_particle\_start\_idx}(\textit{grid\_cell\_array}[j])$  \Comment{Gets start index in \textit{particle\_array} of cell $j$}
	\State $\textit{end\_idx} \gets \text{get\_particle\_end\_idx}(\textit{grid\_cell\_array}[j])$ \Comment{Gets end index in \textit{particle\_array} of cell $j$}
	\State $\textit{m\_occ\_accum} \gets \text{subtract}( \textit{weight\_array\_accum}, \textit{start\_idx}, \textit{end\_idx})$\Comment{Calculate accumulated unnormalized updated particle weight of cell $j$}
	\State $\textit{rho\_p} \gets \text{get\_pers\_occ\_mass}(\textit{grid\_cell\_array}[j])$ \Comment{Gets persistent part of posterior occupancy mass of cell $j$}
	\State $ \textit{mu\_A} \gets \text{calc\_norm\_assoc}(\textit{m\_occ\_accum},\textit{rho\_p})$ \Comment{Calculates normalization component for the case of an associated measurement \eqref{eq:DSnormalizationfactor}}
	\State $ \textit{mu\_UA} \gets \text{calc\_norm\_unassoc}(\textit{grid\_cell\_array}[j])$ \Comment{Calculates normalization component for the case of an unassociated measurement \eqref{eq:DSnormalizationfactor2}}
	\State $\text{set\_normalization\_components}(\textit{grid\_cell\_array}, j, \textit{mu\_A}, \textit{mu\_UA})$ \Comment{Stores $\textit{mu\_A}$ and $\textit{mu\_UA}$ as normalization components in grid cell $j$}
\EndFor
\For {$i \in \{0,\dotsc,\text{length}(\textit{particle\_array})-1 \}$} \Comment{Parallel for loop over all persistent particles}
	\State $\textit{weight\_array}[i] \gets \text{normalize}(\textit{particle\_array}[i], \textit{grid\_cell\_array})$ \Comment{Normalizes particle weights \eqref{eq:DSnormalization}}
\EndFor
\end{algorithmic}
\end{algorithm*}

\begin{algorithm*}[p]
\caption{\small Initialization of New Particles}\label{al:new_particles}
\begin{algorithmic}[1]\footnotesize
\State $\textit{birth\_particle\_array}$  \Comment{This array will be the storage for new-born particles for this recursion (constant size $\numParticlesNewTotal$)}
\State $\textit{particle\_orders\_array\_accum} \gets \text{accumulate}(\textit{born\_masses\_array})$   \Comment{Accumulates mass part of new-born object of each cell}
\State $\text{normalize\_particle\_orders}(\textit{particle\_orders\_array\_accum}, \numParticlesNewTotal)$  \Comment{Normalizes the particle orders to a total number of $\numParticlesNewTotal$}
\For {$j \in \{0,\dotsc,\text{length}(\textit{grid\_cell\_array})-1 \}$ } \Comment{Parallel for loop over all grid cells}
	\State $\textit{start\_idx} \gets \text{calc\_start\_idx}(\textit{particle\_orders\_array\_accum},j)$ \Comment{Calculates first index in $\textit{birth\_particle\_array}$ of cell $j$}
	\State $\textit{end\_idx} \gets \text{calc\_end\_idx}(\textit{particle\_orders\_array\_accum},j)$ \Comment{Calculates last index in $\textit{birth\_particle\_array}$ of cell $j$}
	\State $\textit{num\_new\_particles} \gets \textit{end\_idx} - \textit{start\_idx} + 1$ \Comment{Stores number of new particles for cell $j$}
	\State $\textit{p\_A} \gets \text{get\_assoc\_probability}(\textit{meas\_cell\_array}[j])$ \Comment{Reads association probability to spatial measurement of cell $j$}
	\State $\textit{nu\_A} \gets \text{calc\_num\_assoc}(\textit{num\_new\_particles},p\_A)$ \Comment{Calculates number of new associated particles \eqref{eq:assoc_relation} and \eqref{eq:assoc_sum}}
	\State $\textit{nu\_UA} \gets \textit{num\_new\_particles} - \textit{nu\_A}$ \Comment{Calculates number of new unassociated particles}
	\State $\textit{w\_A} \gets \text{calc\_weight\_assoc}(\textit{nu\_A,p\_A}, \textit{born\_masses\_array}[j]) $ \Comment{Calculates weight of an associated new particle \eqref{eq:weights_associated}}
	\State $\textit{w\_UA} \gets \text{calc\_weight\_unassoc}(\textit{nu\_UA,p\_A}, \textit{born\_masses\_array}[j])$ \Comment{Calculates weight of an unassociated new particle \eqref{eq:weights_unassociated}}
	\State $\text{store\_weights}(\textit{w\_A}, \textit{w\_UA}, \textit{grid\_cell\_array}, j)$ \Comment{Stores weights for new particles in cell $j$}
	\For {$i \in \{\textit{start\_idx} \dotsc \textit{start\_idx} + \textit{nu\_A} \}$ } \Comment{For loop over new associated particles of cell $j$}
		\State $\text{set\_cell\_idx\_A}(\textit{birth\_particle\_array}, i, j)$ \Comment{Sets $j$ as grid cell index of new particle $i$ with flag for associated}
	\EndFor
	\For {$i \in \{ \textit{start\_idx} + \textit{nu\_A} + 1 \dotsc  \textit{end\_idx} \}$ } \Comment{For loop over new unassociated particles of cell $j$}
		\State $\text{set\_cell\_idx\_UA}(\textit{birth\_particle\_array}, i, j)$ \Comment{Sets $j$ as grid cell index of new particle $i$ with flag for unassociated}
	\EndFor
\EndFor
\For {$i \in \{0,\dotsc,\text{length}(\textit{birth\_particle\_array})-1 \}$} \Comment{Parallel for loop over all new-born particles}
		\State $\text{initialize\_new\_particle}(\textit{birth\_particle\_array}, i, \textit{grid\_cell\_array})$ \Comment{Initializes new-born particle $i$ according to \eqref{eq:DSbirth_particle_samplingA} or \eqref{eq:DSbirth_particle_sampling}}
\EndFor
\end{algorithmic}
\end{algorithm*}

\begin{algorithm*}[t]
\caption{\small Statistical Moments of Grid Cells}\label{al:statistical_moments}
\begin{algorithmic}[1]\footnotesize
\State $\textit{vel\_x\_array}$  \Comment{Separate storage for the velocity in $\text{x}$-direction of particles (constant size $\numParticles$)}
\State $\textit{vel\_y\_array}$  \Comment{Separate storage for the velocity in $\text{y}$-direction of particles (constant size $\numParticles$)}
\State $\textit{vel\_x\_squared\_array}$  \Comment{Separate storage for the squared velocity in $\text{x}$-direction of particles (constant size $\numParticles$)}
\State $\textit{vel\_y\_squared\_array}$  \Comment{Separate storage for the squared velocity in $\text{y}$-direction of particles (constant size $\numParticles$)}
\State $\textit{vel\_xy\_array}$  \Comment{Separate storage for the multiplied velocities in both directions of particles (constant size $\numParticles$)}

\For {$i \in \{0,\dotsc,\text{length}(\textit{particle\_array})-1 \}$ } \Comment{Parallel for loop over all persistent particles}
	\State $\textit{w} \gets \textit{weight\_array}[i]$ \Comment{Stores the updated, normalized weight of the persistent particle $i$}
	\State $\textit{vel\_x} \gets   \text{get\_vel\_x}(\textit{particle\_array}[i])$ \Comment{Stores $\text{x}$-velocity of particle $i$}
	\State $\textit{vel\_y} \gets   \text{get\_vel\_y}(\textit{particle\_array}[i])$ \Comment{Stores $\text{y}$-velocity of particle $i$}

	\State $\textit{vel\_x\_array}[i] \gets \textit{w} * \textit{vel\_x} $ \Comment{Stores weighted $\text{x}$-velocity of particle $i$}
	\State $\textit{vel\_y\_array}[i] \gets \textit{w} * \textit{vel\_y} $ \Comment{Stores weighted $\text{y}$-velocity of particle $i$}
	\State $\textit{vel\_x\_squared\_array}[i] \gets \textit{w} * \textit{vel\_x} * \textit{vel\_x} $ \Comment{Stores weighted squared $\text{x}$-velocity of particle $i$}
	\State $\textit{vel\_y\_squared\_array}[i] \gets \textit{w} * \textit{vel\_y} * \textit{vel\_y} $ \Comment{Stores weighted squared $\text{y}$-velocity of particle $i$}
	\State $\textit{vel\_xy\_array}[i] \gets \textit{w} * \textit{vel\_x} * \textit{vel\_y} $ \Comment{Stores weighted product of $\text{x}$- and $\text{y}$-velocity of particle $i$}
\EndFor

\State $\textit{vel\_x\_array\_accum} \gets \text{accumulate}(\textit{vel\_x\_array})$   \Comment{Accumulates velocities in $\text{x}$-direction}
\State $\textit{vel\_y\_array\_accum} \gets \text{accumulate}(\textit{vel\_y\_array})$   \Comment{Accumulates velocities in $\text{y}$-direction}
\State $\textit{vel\_x\_squared\_array\_accum} \gets \text{accumulate}(\textit{vel\_x\_squared\_array})$ \Comment{Accumulates squared velocities in $\text{x}$-direction}
\State $\textit{vel\_y\_squared\_array\_accum} \gets \text{accumulate}(\textit{vel\_y\_squared\_array})$ \Comment{Accumulates squared velocities in $\text{y}$-direction}
\State $\textit{vel\_xy\_array\_accum} \gets \text{accumulate}(\textit{vel\_xy\_array})$ \Comment{Accumulates product of velocities in $\text{x}$- and $\text{y}$-direction}

\For {$j \in \{0,\dotsc,\text{length}(\textit{grid\_cell\_array})-1 \}$} \Comment{Parallel for loop over all grid cells}
	\State $ \textit{rho\_p} \gets \text{get\_pers\_occ\_mass}(\textit{grid\_cell\_array}, j)$ \Comment{Gets persistent part of posterior occupancy mass in grid cell $j$}
	\State $\textit{start\_idx} \gets \text{get\_particle\_start\_idx}(\textit{grid\_cell\_array}[j])$  \Comment{Gets start index in \textit{particle\_array} of cell $j$}
	\State $\textit{end\_idx} \gets \text{get\_particle\_end\_idx}(\textit{grid\_cell\_array}[j])$  \Comment{Gets end index in \textit{particle\_array} of cell $j$}
	\State $\textit{mean\_x\_vel} \gets \text{calc\_mean}(\textit{vel\_x\_array\_accum}, \textit{start\_idx}, \textit{end\_idx}, \textit{rho\_p})$  \Comment{Applies \eqref{eq:mean_x}}
	\State $\textit{mean\_y\_vel} \gets \text{calc\_mean}(\textit{vel\_y\_array\_accum}, \textit{start\_idx}, \textit{end\_idx}, \textit{rho\_p})$  \Comment{Applies \eqref{eq:mean_x}}
	\State $\textit{var\_x\_vel} \gets \text{calc\_variance}(\textit{vel\_x\_squared\_array\_accum}, \textit{start\_idx}, \textit{end\_idx}, \textit{rho\_p}, \textit{mean\_x\_vel})$  \Comment{Applies \eqref{eq:var_x}}
	\State $\textit{var\_y\_vel} \gets \text{calc\_variance}(\textit{vel\_y\_squared\_array\_accum}, \textit{start\_idx}, \textit{end\_idx}, \textit{rho\_p}, \textit{mean\_y\_vel})$  \Comment{Applies \eqref{eq:var_x}}
	\State $\textit{covar\_xy\_vel} \gets \text{calc\_covariance}(\textit{vel\_xy\_array\_accum}, \textit{start\_idx}, \textit{end\_idx}, \textit{rho\_p}, \textit{mean\_x\_vel}, \textit{mean\_y\_vel})$  \Comment{Applies \eqref{eq:covar_xy}}
	\State $\text{store}(\textit{grid\_cell\_array}, j, \textit{mean\_x\_vel}, \textit{mean\_y\_vel}, \textit{var\_x\_vel}, \textit{var\_y\_vel}, \textit{covar\_xy\_vel})$  \Comment{Store stochastic moments in grid cell $j$}
\EndFor
\end{algorithmic}
\end{algorithm*}

\begin{algorithm*}[t]
\caption{\small Resampling}\label{al:resampling}
\begin{algorithmic}[1]\footnotesize
\State $\textit{rand\_array}$   \Comment{Array with sorted, equally distributed random numbers (constant size $\nu$)}
\State $\textit{idx\_array\_resampled}$   \Comment{Array with indices of resampled particles (constant size $\nu$)}
\State $\textit{particle\_array\_next}$   \Comment{Particle array for the next time step (constant size $\nu$)}
\State $\textit{joint\_weight\_array\_accum} \gets \text{accumulate}(\textit{weight\_array}, \textit{birth\_weight\_array})$   \Comment{Accumulates normalized particle weights}
\State $\textit{idx\_array\_resampled} \gets \text{calc\_resampled\_indeces}(\textit{joint\_weight\_array\_accum}, \textit{rand\_array})$   
\Comment{Calculates resampled particle indices}

\For {$i \in \{0,\dotsc,\text{length}(\textit{particle\_array})-1 \}$ } \Comment{Parallel for loop over all persistent particles of the next time step}
\State $\textit{particle\_array\_next}[i] \gets \text{copy\_particle}(\textit{particle\_array}, \textit{birth\_particle\_array}, \textit{idx\_array\_resampled}[i])$   \Comment{Copy resampled particle}

\EndFor
\end{algorithmic}
\end{algorithm*}

\subsubsection{Initialization of New Particles}
Algorithm \ref{al:new_particles} depicts pseudo code for the particle initialization. New particles are stored in the array $\textit{birth\_particle\_array}$. The total number of new particles for all grid cells $\numParticlesNewTotal$ remains constant over time, which is feasible for many real-time applications. The goal of this algorithm is to initialize for each grid cell a certain number of new particles, which is proportional to the new-born part of its updated occupancy mass $\massbornkpcell$. Therefore, the algorithm accumulates new-born occupancy masses of all grid cells in the array $\textit{particle\_orders\_array}$. Then it normalizes each value of the array to sum up to the discrete number $\numParticlesNewTotal$. The normalized array then serves as a lookup table to be used by each grid cell to find out its first and last corresponding index in the array $\textit{birth\_particle\_array}$, as well as the individual number of new particles $\numParticlesNewCell$ assigned to grid cell $c$ in this time step. 

Each grid cell splits this number into the number $\numParticlesAssCellp$ of new particles which are associated to the  spatial measurement $\likecell$ and the number $\numParticlesAssNCellp$ of new particles which are not associated.
The relation is given by
\begin{align}
\frac{\numParticlesAssCellp}{\numParticlesAssNCellp} = \frac{\pA}{1-\pA},
\label{eq:assoc_relation}
\end{align}
and
\begin{align}
\numParticlesAssCellp + \numParticlesAssNCellp  = \numParticlesNewCell.
\label{eq:assoc_sum}
\end{align}
Each grid cell also calculates and stores the weights for associated \eqref{eq:weights_associated} and unassociated \eqref{eq:weights_unassociated} new particles.
As a next step, each grid cell iterates over all its assigned new particles and defines it as an associated or unassociated particle, respectively, and sets the particle grid cell index. 
Finally, each particle initializes itself with a random initial state within its grid cell. Again, enough random numbers should be sampled in a separate step in advance so the initialization step just needs to look them up. This time, each particle uses the grid map as a lookup table for its initial position and weight. 

\subsubsection{Statistical Moments of Grid Cells}
Algorithm \ref{al:statistical_moments} calculates the first two statistical moments of the two-dimensional velocity $[v_\text{x} \, v_\text{y}]\transp$ in a grid cell, considering all updated, normalized, persistent particles. If grid cell $c$ contains a certain number of particles, the mean velocity component in $\text{x}$-direction $\vxmean$ can be approximated by
\begin{align}
\vxmean \approx \frac{1}{\massperskpcell} 
\sum_{i=1}^{\numParticlesPersCellp} \weightperspcell \cdot \partvx,
\label{eq:mean_x} 
\end{align}
and analogously for the component in $\text{y}$-direction.
The symbol $\partvx$ denotes the velocity $\text{x}$-component of a posterior persistent particle $\partperspcell$ in grid cell $c$.

Recall that $\massperskpcell$ is the persistent part of the posterior occupancy mass, which equals the sum of updated, normalized weights $\weightperspcell$ of persistent particles in grid cell $c$:
\begin{align}
\massperskpcell = 
\sum_{i=1}^{\numParticlesPersCellp} \weightperspcell.
\label{eq:occupancyProbPersParticles}
\end{align}

The variance of the velocity component in $\text{x}$-direction can be approximated by
\begin{align}
\varx \approx \frac{1}{\massperskpcell} 
\sum_{i=1}^{\numParticlesPersCellp} \weightperspcell \cdot \left({\partvx}\right)^2 \,\,\, - \,\,\, \left(\vxmean\right)^2 ,
\label{eq:var_x} 
\end{align}
and analogously for the component in $\text{y}$-direction.
The covariance of the velocity components in $\text{x}$- and $\text{y}$-direction can by approximated by 
\begin{align}
\covarxy\approx\ &\frac{1}{\massperskpcell} 
\sum_{i=1}^{\numParticlesPersCellp} \weightperspcell \cdot
\partvx \cdot \partvy  \nonumber \\[.5em]
&- \vxmean \cdot \vymean.
\label{eq:covar_xy} 
\end{align}
Since particles are sorted by their grid cell index, the calculation of all sums in equations \eqref{eq:mean_x}, \eqref{eq:occupancyProbPersParticles},  \eqref{eq:var_x}, and \eqref{eq:covar_xy} can be realized by parallel accumulation of the according values. A parallel \textbf{\footnotesize for} loop over all grid cells then only subtracts the corresponding accumulated values. Again, the computational complexity is constant for each cell, independent of the individual number of particles in the cell at time step $k+1$. This is optimal with respect to load balancing between threads.

\subsubsection{Resampling}
Particles are resampled according to Algorithm \ref{al:resampling} to avoid degeneration. The resampling step accumulates the normalized weights of persistent and new-born particles in the array \textit{joint\_weight\_array\_accum}. It draws $\nu$ sorted random numbers which are equally distributed between $0$ and the sum of all particles and stores the random numbers in the array \textit{rand\_array}. Each random number falls into a certain interval of accumulated weights, which corresponds to a certain particle index. For each random number in \textit{rand\_array}, the corresponding particle is chosen and copied into the particle array for the next time step \textit{particle\_array\_next}. 

\section{Evaluation}
\label{sec:evaluation}

This section evaluates the Dempster-Shafer approximation of the PHD/MIB filter (DS-PHD/MIB filter) with real-world sensor data. The goal is to investigate if the DS-PHD/MIB filter performs as expected in different scenarios. A focus lies on the effect of the birth probability, which will be varied in all experiments. 

A test vehicle equipped with laser and radar sensors is used for recording measurement data. In a first experiment, an object approaches the ego vehicle with varying speed. The evaluation examines the speed estimation performance and consistency of the DS-PHD/MIB filter. In a second experiment, the ego vehicle follows a dynamic object and the evaluation investigates the ability of the DS-PHD/MIB filter to separate dynamic and static obstacles in the vehicle's environment. The evaluation also determines the effect of fusing radar and laser data in comparison to using laser data only. Finally, the computation time of the parallel implementation with varying numbers of particles is analyzed.

\begin{table}[t]
  \caption{Overview of the experimental parameters of the DS-PHD/MIB filter. Here, SD is an abbreviation for standard deviation.}
  \label{tab:params}
  \centering
  \renewcommand{\arraystretch}{1.15}
  \begin{tabular}{lcr}\toprule
    Parameter       & Symbol  & Value \\ \midrule
    Grid map edge size    & -     & $120$ m \\
    Grid cell edge size   & -     & $0.1$ m \\
    Number of consistent particles  & $\numParticles$     & $2 \cdot 10^6$\\
    Number of new-born particles per step   & $\numParticlesNewTotal$     & $2 \cdot 10^5$ \\
    Persistence probability   & $\pS$     & $0.99$ \\
    SD velocity new-born particles  & $\sigma_{\text{B,v}}$   & $4$ m/s \\
    SD process noise position & ${\sigma}_{\text{p}}$   & $0.02 \frac{\text{m}}{T \!/ \!\text{s}}$  \\
    SD process noise velocity & ${\sigma}_{\text{v}}$   & $0.8 \frac{\text{m/s}}{T \!/ \!\text{s}}$ \\
    Birth probability     & $\pB$     & $0.005 $ ... $ 0.1$  \\ \bottomrule
  \end{tabular}
\end{table}

\subsection{Experiment Configuration}
The test vehicle is equipped with a Valeo four-layer laser scanner with an opening angle of $120$~degrees in the front bumper. Additionally, two short range Delphi single beam mono pulse radars facing to the front left and front right sides cover a similar area. The vehicle speed and yaw rate are available via CAN messages, so the ego movement of the test vehicle can be compensated in the grid map.

The grid map covers an area of $120$~m by $120$~m with the test vehicle in the center. Each grid cell measures $10$~cm by $10$~cm. Table \ref{tab:params} shows the parameter set of the DS-PHD/MIB filter. The parallel implementation is tested on an Nvidia GTX980 GPU, supported by a single core of an Intel i7 processor.

\subsection{Velocity Estimation of a Moving Object}
\begin{figure}[t]
  \centering
  \includegraphics[width=0.9\columnwidth]{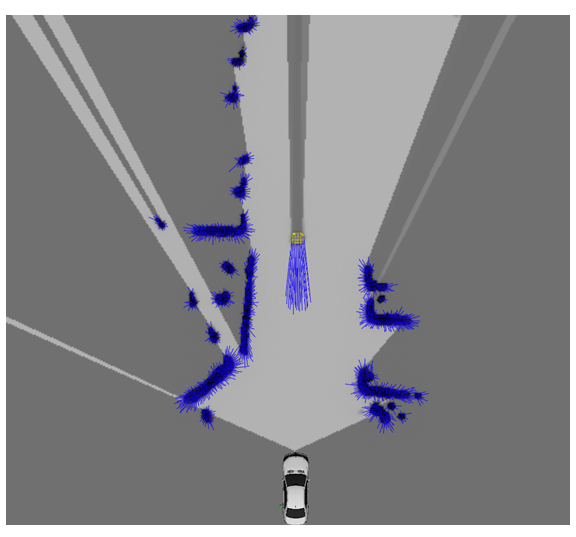}
	\caption{Velocity estimation test scenario: A Segway approaches the test vehicle. The estimated mean velocity of every grid cell is visualized as a blue vector.}
	\label{fig:approaching_scene}
\end{figure}

In this test scenario, a Segway approaches the test vehicle starting from a distance of ca. $50$~m, see Fig. \ref{fig:approaching_scene}. The Segway accelerates slowly, drives with almost constant speed and then decelerates strongly. The set of grid cells representing the Segway is denoted $S$. The experiment evaluates the mean velocity $\vxmeanS$ of these grid cells:
\begin{align}
\vxmeanS = \sum_{c \in S} \frac{1}{|S|} \vxmean.
\label{eq:mean_cluster} 
\end{align}

The $\text{x}$-component of the estimated velocity is aligned to the longitudinal axis of the test vehicle. 

\begin{figure}[t]
  \centering
	  \input{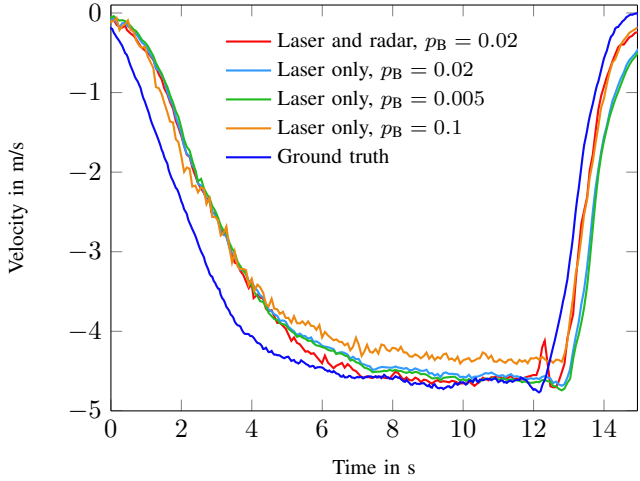}
	  	\caption{Velocity estimation results under variation of the birth probability $\pB$. The Segway reaches the short range radar field of view after a time of ca. $5$~s. Fusion with radar further improves the velocity estimation.}
	\label{fig:velocity_results}
\end{figure}

Figure \ref{fig:velocity_results} shows the results of the experiment. In the beginning the Segway accelerates slowly. Since the process model assumes constant velocity, the estimated velocity is delayed during the acceleration phase. During the constant velocity phase, the estimation converges closely to the real velocity for birth probabilities $\pB = 0.005$ and $\pB = 0.02$. Choosing a process model with a higher birth probability of $\pB = 0.1$ results in a persistent bias of the estimated velocity. The reason is that the mean of the birth distribution is zero, so new-born particles generally distort the velocity estimation towards zero. The results show that choosing an appropriate birth probability is important for the velocity estimation performance of the filter.

Fusion of radar data which contains Doppler measurements further improves the velocity estimation. In this realization, the Segway reaches the radar field of view at a time of ca. $4$~s. Before that point in time, the Segway is outside the radar range and only rarely detected by radar. Radar Doppler measurements reduce even small remaining bias effects and lead to a much faster convergence during the strong deceleration phase in the end. The small peak after $12$~s is assumed to be caused by a movement of the person riding the Segway, commonly referred to as micro-Doppler.

\subsection{Consistency of the DS-MIB/PHD filter}
This section evaluates the consistency of the DS-MIB/PHD filter in dependency of the birth probability $\pB$, focusing on the $\text{x}$-component of the estimated velocity. The DS-MIB/PHD filter provides for each grid cell the estimation variance $\varx$. The experiment considers the combined distribution of the set $S$ of all grid cells representing the Segway as a Gaussian mixture. Hence the combined variance $\varxS$ of the Segway is given by
\begin{align}
\varxS = \sum_{c \in S} \frac{1}{|S|} \left(\varx  + \vxmeanSqared \right) - \vxmeanSSquared  . 
\label{eq:gaussian_mixture_x}
\end{align}
This corresponds to the variance of all particles representing the Segway. 

\begin{figure}[t]
  \centering
    \input{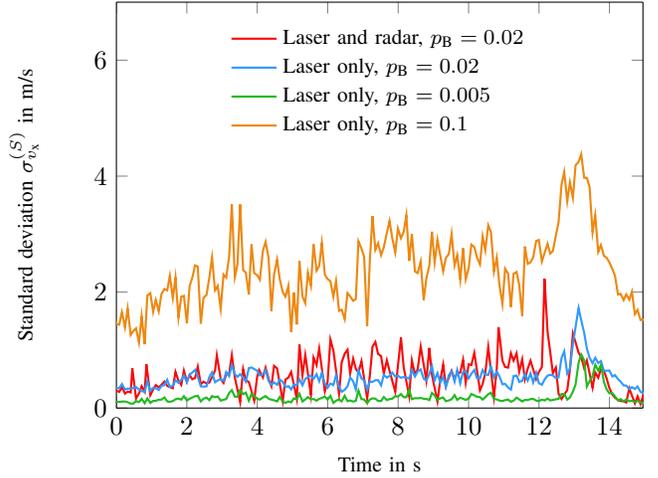}
    \caption{Estimation standard deviation $\stddevS$ of the Segway's velocity $\text{x}$-component provided by the DS-PHD/MIB filter under variation of the birth probability $\pB$.}
  \label{fig:velocity_uncertainty}
\end{figure}

Figure \ref{fig:velocity_uncertainty} shows the standard deviation $\stddevS$ of the velocity $\text{x}$-component of the Segway provided by the DS-MIB/PHD filter. The test scenario is the same as used for the velocity estimation. The uncertainty increases with the birth probability $\pB$ due to the high number of new-born particles in the dynamic object. 

\begin{figure}[t]
  \centering
	  \input{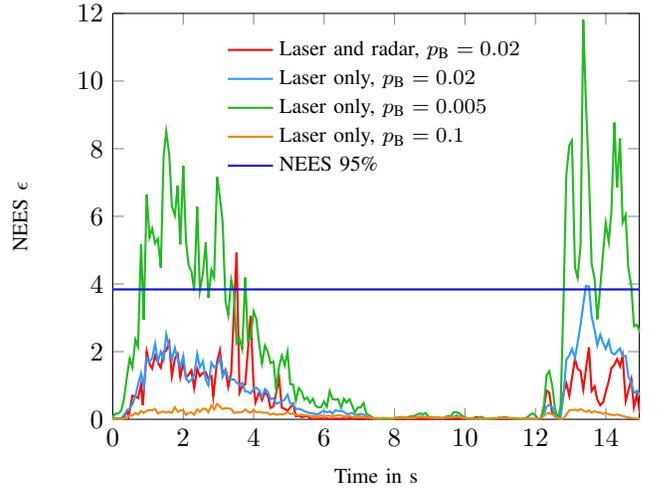}
  	\caption{Normalized estimation error squared (NEES) $\epsilon$ of the estimated velocity $\text{x}$-component under variation of the birth probability $\pB$.}	\label{fig:velocity_consistency}
\end{figure}

To evaluate the consistency of the DS-PHD/MIB filter, the experiment calculates the normalized estimation error squared (NEES) $\epsilon$, given by \cite{Bar-Shalom2001}:
\begin{align}
\epsilon = \frac{\left( \vxmeanS - v^{(S)}_\text{x}  \right)^2}{\varxS},
\label{eq:NEES_x}
\end{align}
where $v^{(S)}_{\text{x}}$ is the true velocity $\text{x}$-component of the Segway.

Figure \ref{fig:velocity_consistency} shows the result of the consistency test and compares the NEES $\epsilon$ to the $95$\% level \cite{Bar-Shalom2001}. 
The applied process model \eqref{eq:process} assumes constant velocity and does not model acceleration maneuvers. To compensate this, the filter designer can choose a higher velocity process noise than expected during constant velocity maneuvers as a trade-off between both modes. The consistency check shows why the birth probability $\pB$ should not be chosen too small. Especially during acceleration maneuvers, the result can become inconsistent, because the filter underestimates the uncertainty of the estimation result, which happens during the deceleration phase with a birth probability of $\pB = 0.005$. 

\subsection{Separation of Moving and Stationary Obstacles}
\begin{figure}[t]
  \centering
  \includegraphics[width=0.9\columnwidth]{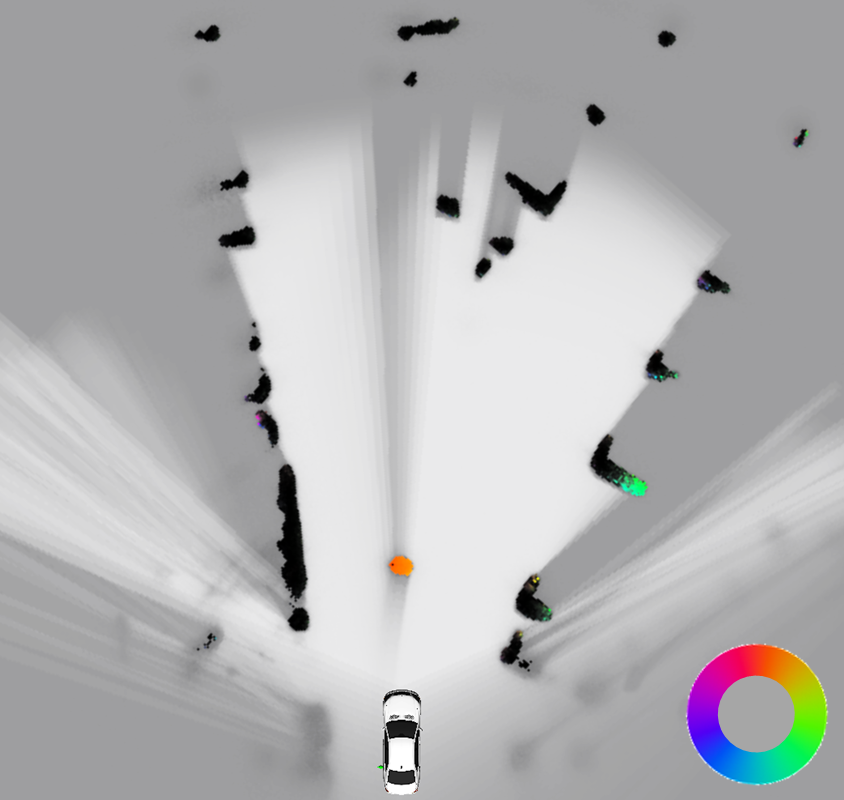}
	\caption{Test scenario for separation of static and dynamic grid cells. The color code represents the direction of movement, the color saturation is determined by the Mahalanobis distance between the velocity distribution and the velocity $v=0$ in a grid cell.}
	\label{fig:follow_color}
\end{figure}

In the second test scenario, the Segway drives along between parked vehicles with the ego vehicle following behind. Except for the Segway, the environment is static. In a manual post-processing step, all grid cells were labeled as dynamic or static. An example situation of the test is depicted in Fig. \ref{fig:follow_color}.

The evaluation uses the DS-PHD/MIB filter as a classifier to separate grid cells into dynamic or static. The criterion for the assignment is the Mahalanobis distance $m$ between the estimated two-dimensional probability density $p(v) : \mathbb{R}^2  \rightarrow \mathbb{R}$ of the velocity distribution in a cell and the velocity $v=0$. The density is approximated from the particle representation as a Gaussian distribution with mean $\overline{v}$ and covariance matrix $P$ as calculated in \eqref{eq:mean_x}, \eqref{eq:var_x}, \eqref{eq:covar_xy}. Then the Mahalanobis distance is given by 
\begin{align}
m = \overline{v} P ^{-1} \overline{v}^T.
\label{eq:mahadist} 
\end{align}
Applying different threshold values $\tau_m$, the evaluation classifies grid cells as a static detection if $m<\tau_m$ or as a dynamic detection if $m \geq \tau_m$. 

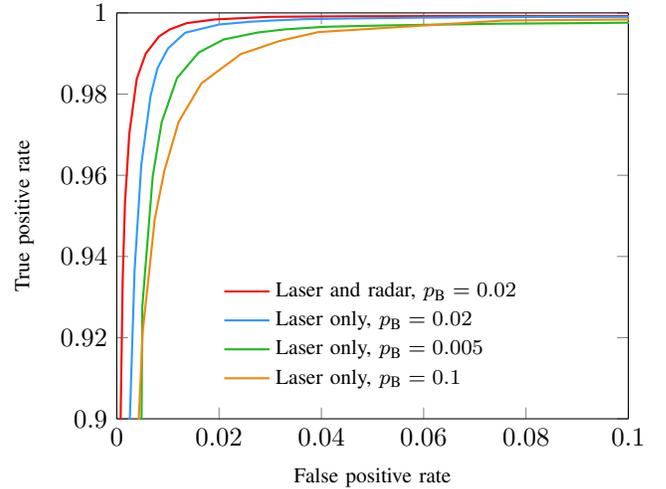
\begin{figure}[t]
  \centering
\begin{tikzpicture}[every node/.style={font=\footnotesize}]
\begin{axis}[%
width=68mm,
height=54mm,
scale only axis,
separate axis lines,
every outer x axis line/.append style={black},
every x tick label/.append style={font=\color{black}},
xmin=0,
xmax=0.1,
xlabel={False positive rate},
every outer y axis line/.append style={black},
every y tick label/.append style={font=\color{black}},
tick label style={/pgf/number format/fixed},
ymin=0.9,
ymax=1,
ylabel={True positive rate},
axis background/.style={fill=white},
legend style={at={(0.5,0.05)},anchor=south,legend cell align=left,align=left,draw=none,fill=none},
cycle list name=theplotstyles  % See root.tex for definition
]
\addplot
  table[row sep=crcr]{%
0.991354394805232	0.99923329953256\\
0.684023108157814	0.99923329953256\\
0.358673132078556	0.99923329953256\\
0.269403808566751	0.99923329953256\\
0.176767790901463	0.99923329953256\\
0.120402830100227	0.99923329953256\\
0.105934507653435	0.99923329953256\\
0.089189507569292	0.99923329953256\\
0.0696297934642283	0.99923329953256\\
0.0562500450767748	0.999208567259417\\
0.0468106078272514	0.999183834986274\\
0.0396454441204259	0.999134370439987\\
0.0298203900883986	0.999035441347414\\
0.019139478359542	0.99839240224569\\
0.0136408330668795	0.997477308139391\\
0.0103736684320735	0.995943907204511\\
0.00827056642874143	0.994212648084485\\
0.00567270174561313	0.990008161650137\\
0.00390569217491231	0.983775628818045\\
0.002449772452441	0.970222343135558\\
0.00160256949636621	0.953206539213019\\
0.00115685034751188	0.934558405263028\\
0.000722670852996824	0.89760838918705\\
4.85627120110973e-05	0.00507011599436104\\
};
\addlegendentry{Laser and radar, $\pB=0.02$};

\addplot
  table[row sep=crcr]{%
0.993741470814742	0.999158915495745\\
0.69268782322651	0.999158915495745\\
0.312902770307241	0.999158915495745\\
0.21741166109744	0.999158915495745\\
0.12993159930178	0.999084702157134\\
0.0831505567833557	0.999010488818524\\
0.0720520488159703	0.998936275479913\\
0.0596929168164645	0.998837324361765\\
0.0457600949787775	0.99863942212547\\
0.036719274727786	0.998466257668712\\
0.0305699005342556	0.998144666534732\\
0.0260369491969444	0.997823075400752\\
0.0200143805895484	0.997155155353255\\
0.0133819374964583	0.995126657431229\\
0.0100120674327025	0.991242826043934\\
0.00798271356522499	0.986468434593311\\
0.00661107658407117	0.97959133188205\\
0.00479678127476309	0.962670690678805\\
0.00348599239538075	0.936844448842272\\
0.00227506872119586	0.88900158321789\\
0.00150912582179774	0.825202849792203\\
0.00107575728017346	0.763259449831783\\
0.000680476863322158	0.652904215317633\\
1.81150837334909e-05	0.00274589352859687\\
};
\addlegendentry{Laser only, $\pB=0.02$};

\addplot
  table[row sep=crcr]{%
0.993150534811144	0.998217990967899\\
0.828018620096446	0.998217990967899\\
0.545881755141348	0.998217990967899\\
0.441470463386694	0.998217990967899\\
0.316291600336189	0.998144757720005\\
0.230817986186561	0.99807152447211\\
0.207446696686287	0.997998291224216\\
0.179746910181339	0.997925057976321\\
0.146535972328393	0.997876235811058\\
0.123223278788397	0.997778591480532\\
0.106021562418788	0.997680947150006\\
0.0928567051446806	0.997510069571586\\
0.0743699033256115	0.997339191993165\\
0.0527032642422849	0.996924203588429\\
0.0406298126405353	0.996582448431588\\
0.0328607039834965	0.995947760283169\\
0.0276304554069087	0.995191016721592\\
0.0208834213241882	0.993457829854754\\
0.0159998282377958	0.990235566947394\\
0.0117670528781575	0.983937507628463\\
0.00878492106766313	0.973098986940071\\
0.00702525306969028	0.959721713658001\\
0.0049931809510348	0.927694373245453\\
5.54648784312743e-05	0.00285609666788722\\
};
\addlegendentry{Laser only, $\pB=0.005$};

\addplot
  table[row sep=crcr]{%
0.99367576147036	0.99922324064005\\
0.670200890286541	0.99922324064005\\
0.261197295666341	0.999197348661385\\
0.15936866123814	0.99901610481073\\
0.0756702612868669	0.998135777536119\\
0.0394364077751896	0.995287659882968\\
0.0318781340322958	0.993112733675107\\
0.0242287535834878	0.989850344363316\\
0.0165409854776934	0.982626482315779\\
0.0120885291027064	0.973150018124385\\
0.00927855261130757	0.960980788151831\\
0.00742519653200788	0.949174045880586\\
0.00509941635406318	0.921806224431671\\
0.00292411579150486	0.862591269224794\\
0.00187741234434445	0.803091502252602\\
0.00131848805862358	0.743514059344415\\
0.00100575661304167	0.690953342654446\\
0.000625462891163827	0.592123660090104\\
0.000394625113819403	0.494795712288333\\
0.000238515308741688	0.3770907772772\\
0.000155086134224746	0.281316348195329\\
0.000108509110414674	0.214256123452954\\
7.01214534283505e-05	0.134042773548755\\
3.07101255890586e-06	0.000543731551965201\\
};
\addlegendentry{Laser only, $\pB=0.1$};

\end{axis}
\end{tikzpicture}%
	  	\caption{Detection of dynamic grid cells with the DS-PHD/MIB filter under variation of the birth probability $\pB$. }
	\label{fig:ROC}
\end{figure}

Figure \ref{fig:ROC} shows the receiver operating characteristic (ROC) curve of the assignment. It shows that the DS-PHD/MIB filter is able to achieve a true positive rate of 99\% (ratio of correctly detected dynamic cells to total number of dynamic cells) at a false positive rate of 1\% (ratio of falsely detected dynamic cells to total number of static cells) in the test scenario. The birth probability $\pB$ of the process model has an important influence on the estimation result. Best classification performance is achieved with a birth probability value of $\pB = 0.02$, which also delivered consistent estimation results in the previous experiment.

\begin{figure}[t]
  \centering
  \includegraphics[width=\columnwidth]{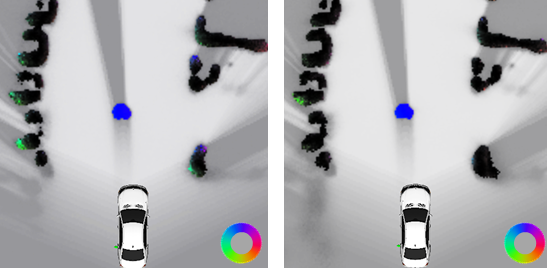}
	\caption{Left: Laser only, false positive movements occur on both sides. Right: Fusion of laser and radar, false positive movements are reduced. The color code represents the direction of movement.}
	\label{fig:laser_radar}
\end{figure}

Again, fusion with radar data further improves the overall result. As described in details in \cite{Nuss2015}, Doppler measurements help reduce false positive movement estimation in grid cells. An exemplary visualization of the dynamic grid map with and without radar is given in Fig. \ref{fig:laser_radar}.

\subsection{Computation Time}

\begin{figure}[t]
  \centering
\begin{tikzpicture}[every node/.style={font=\footnotesize}]

\begin{axis}[%
width=70mm,
height=54mm,
scale only axis,
separate axis lines,
every outer x axis line/.append style={black},
every x tick label/.append style={font=\color{black}},
xmin=1000000,
xmax=10000000,
xlabel={Number of particles},
every outer y axis line/.append style={black},
every y tick label/.append style={font=\color{black}},
ymin=0,
ymax=150,
ylabel={Time in ms},
axis background/.style={fill=white},
legend style={at={(0.5,0.96)},anchor=north,legend cell align=left,align=left,draw=none,fill=none},
cycle list name=theplotstyleswithmarkers  % See root.tex for definition
]
\addplot
  table[row sep=crcr]{%
1000000	22.034\\
2000000	31.055\\
3000000	42.079\\
4000000	52.043\\
7000000	86.81\\
10000000	119.152\\
};
\addlegendentry{Complete recursion};

\addplot
  table[row sep=crcr]{%
1000000	8.231\\
2000000	16.442\\
3000000	24.805\\
4000000	32.669\\
7000000	57.625\\
10000000	81.574\\
};
\addlegendentry{Sort and assign particles};

\end{axis}
\end{tikzpicture}%
	  	\caption{Parallel implementation of the DS-PHD/MIB filter: Computation time with varying number of particles and $1.44 \cdot 10^6$ grid cells.}
	\label{fig:timing}
\end{figure}
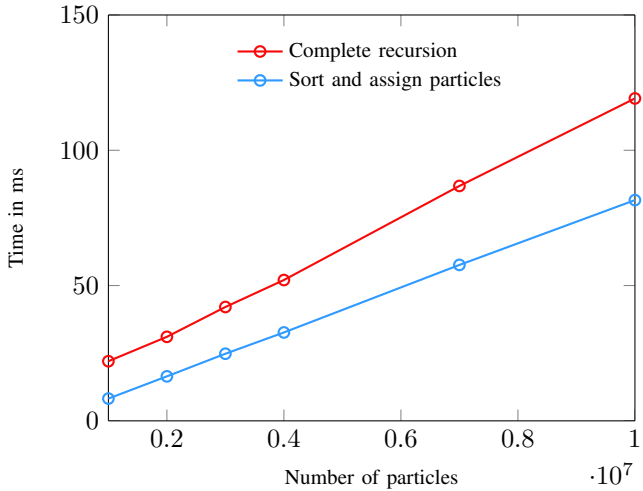

The parallel implementation presented in Sect. \ref{sec:implementation} allows to run the DS-PHD/MIB filter in real-time applications.
Figure \ref{fig:timing} shows the computation time of one recursion of the DS-PHD/MIB filter in dependence on the number of persistent particles $\numParticles$, running on an Nvidia GTX980 GPU.
The number of new-born particles $\numParticlesNewTotal$ is chosen to be 10\% of the number of persistent particles in this experiment.

As discussed in Sect. \ref{sec:implementation}, the time complexity of the sorting step is above linear. However, in the range between ${1 \cdot 10^6}$ and ${10 \cdot 10^6}$ particles, the absolute computation time relates approximately linear to the number of particles. A typical environment perception application with a refresh time of ca. $50$~ms can process more then ${2 \cdot 10^6}$ particles and ${1.44 \cdot 10^6}$ grid cells in each update step, which are also the numbers that have been used during the experiments.

\section{Conclusion and Outlook}
\label{sec:conclusion}
This paper presented the first mathematically rigorous approach for the dynamic state estimation of grid cells for robotic or vehicle environment perception based on random finite sets (RFSs). The PHD/MIB filter approximates the multi-object estimation problem by combining the probability hypothesis density filter and multiple instances of the Bernoulli filter. In contrast to former approaches, the top-down derivation of the proposed PHD/MIB filter facilitates a characterization of the approximation error. Further, the proposed RFS formulation provides an explicit, stochastic birth model for appearing objects as well as a physical meaning for the densities represented by the particles. The validity of the PHD/MIB filter was additionally verified by the proof that the filter corresponds to the well-known binary Bayes filter in case of a static process model. Moreover, an approximate particle realization of the PHD/MIB filter in the Dempster-Shafer domain called DS-PHD/MIB filter was proposed which facilitates a real-time capable implementation for practical applications in robotics or vehicle environment perception since it requires a significantly smaller number of particles. Further, an efficient parallel algorithm suitable for a GPU implementation of the filter was presented as pseudo code.

The quantitative evaluation with real-world sensor data showed that appropriate stochastic models for the system process and for the observation process lead to consistent estimation results. The experiments have confirmed that the DS-PHD/MIB filter provides useful results in regard to velocity estimation of dynamic obstacles and separation of dynamic and static obstacles. 

Further research should investigate possibilities of explicitly modeling the dynamic behavior of free space, which is bound to a naive model in the DS-PHD/MIB approximation. Theoretically, the PHD/MIB filter is able to model possible movements in occluded areas which could lead to useful applications. 
\bibliographystyle{IEEEtrancustomized}
%\clearpage
\bibliography{mrm}

\end{document}